\documentclass[10pt,twocolumn,letterpaper]{article}
\usepackage{wacv} %
\usepackage{booktabs}
\usepackage{multirow}
\usepackage{amsmath}
\usepackage{amssymb}
\usepackage{graphicx}
\usepackage{subcaption}
\usepackage{xcolor}
\usepackage{enumitem}

\usepackage[accsupp]{axessibility}

\usepackage[T1,OT1]{fontenc}
\definecolor{wacvblue}{rgb}{0.21,0.49,0.74}
\usepackage[breaklinks,colorlinks,allcolors=wacvblue,hyperfootnotes=false]{hyperref}
\def\wacvPaperID{278}
\def\confName{WACV}
\def\confYear{2027}
\title{NightIQA: A Comprehensive Dataset and Benchmark for MLLM-Based Nighttime Image Quality Assessment}

\author{
Yiding Tian\textsuperscript{1,2,*,$\dagger$}\quad
Jiaojiao Yi\textsuperscript{2,*}\quad
Jiayuan Lu\textsuperscript{2,*}\quad
Pufan Xu\textsuperscript{3}\\[0.15em]
Jinlin Wu\textsuperscript{4,5}\qquad
Changyuan Wen\textsuperscript{2}\qquad
Hongyuan Yu\textsuperscript{2,$\dagger$,$\ddagger$}\\[0.45em]
{\small\textsuperscript{1}Xiamen University Malaysia\quad
\textsuperscript{2}Xiaomi Corporation}\\[-0.05em]
{\small\textsuperscript{3}Tsinghua University\quad
\textsuperscript{4}Institute of Automation, Chinese Academy of Sciences}\\[-0.05em]
{\small\textsuperscript{5}Centre for Artificial Intelligence and Robotics, HKISI, CAS, Hong Kong}\vspace{-1.0em}
}

\hypersetup{pdftitle={NightIQA: A Comprehensive Dataset and Benchmark for MLLM-Based Nighttime Image Quality Assessment},pdfauthor={Yiding Tian, Jiaojiao Yi, Jiayuan Lu, Pufan Xu, Jinlin Wu, Changyuan Wen, Hongyuan Yu}}
\graphicspath{{./}}
\begin{document}
\AddToShipoutPicture*{\AtPageLowerLeft{\put(54,18){\parbox[b]{504pt}{\fontsize{6.5}{7.5}\selectfont
Accepted to WACV 2027. Author's accepted manuscript.\\
\textcopyright\ 2027 IEEE. Personal use of this material is permitted. Permission from IEEE must be obtained for all other uses, in any current or future media, including reprinting/republishing this material for advertising or promotional purposes, creating new collective works, for resale or redistribution to servers or lists, or reuse of any copyrighted component of this work in other works.
}}}}
\maketitle
\thispagestyle{empty}
\begingroup
\renewcommand{\thefootnote}{*}
\footnotetext{\footnotesize Equal contribution.\quad
\textsuperscript{$\dagger$}Project leaders.\quad
\textsuperscript{$\ddagger$}Corresponding author.}
\endgroup

\begin{abstract}
Nighttime images are inherently susceptible to complex, compound degradations, making accurate quality assessment exceptionally challenging. Score-based IQA methods condense quality into a single number, lacking the multi-dimensional, language-grounded reasoning required for nighttime evaluation. Multimodal Large Language Models (MLLMs) offer a more comprehensive alternative, yet remain constrained by a critical data gap: existing real-world IQA datasets are overwhelmingly daytime-centric and often built from outdated captures, while synthetic alternatives lack nighttime-specific degradation modeling, yielding artifacts far removed from authentic nighttime conditions. To bridge this gap, we present NightIQA, a dataset and benchmark suite purpose-built for MLLM-based nighttime IQA. It comprises three core components: \textbf{NightIQA-Real}, a real-world dataset of 1,769 high-resolution images that we captured ourselves with modern equipment, encompassing six canonical distortion types and structured multi-dimensional annotations; \textbf{NightIQA-Bench}, a benchmark of 1,078 visually grounded multiple-choice questions tailored to MLLM-based nighttime IQA, organized around a four-level evaluation hierarchy spanning Perception, Technical Assessment, Localization, and Reasoning; and \textbf{NightIQA-Syn}, an inherently scalable, scene-aware synthesis pipeline that jointly models global-level and instance-level degradations, with an initial release of 80k high-fidelity images alongside the full generation code. Fine-tuned on NightIQA with Supervised Fine-Tuning (SFT) and Group Relative Policy Optimization (GRPO), Qwen3-VL-8B outperforms open-source IQA methods and achieves higher overall accuracy than leading closed-source models, highlighting the value of domain-specific data and structured supervision.
\end{abstract}

\begin{figure}[!t]
    \centering
    \includegraphics[width=0.89\columnwidth]{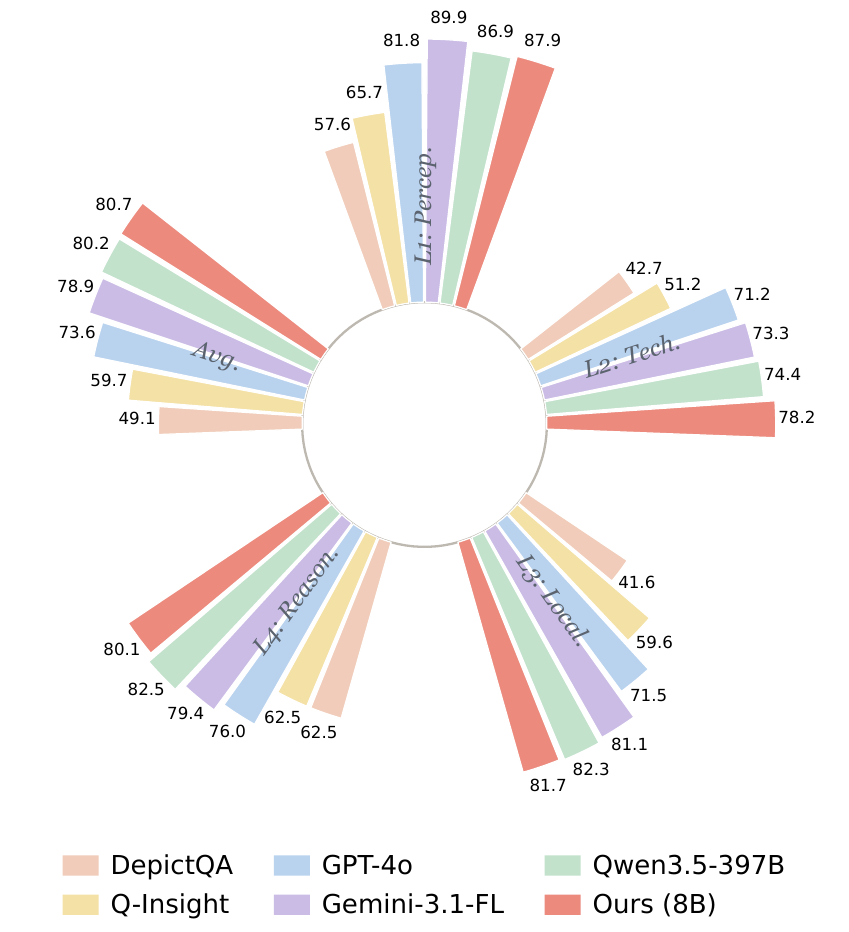}
    \caption{NightIQA-Bench accuracy of six models (L1--L4 and average). Trained on NightIQA, our 8B model (red) leads on overall and L2 accuracy, ahead of larger VLMs and IQA baselines, indicating that domain-specific data, not model scale, is the bottleneck for nighttime IQA.}
    \vspace{-14pt}
    \label{fig:benchmark_petal}
\end{figure}

\section{Introduction}
\label{sec:intro}

Nighttime images are prone to complex, compound degradations that pose unique challenges for Image Quality Assessment (IQA)~\cite{xiang2019blind,zhai2020perceptual}. In such conditions, multiple artifacts---from sensor noise and exposure failures to various forms of motion and optical blur---tend to co-occur and interact within a single capture, producing degradation patterns far more intricate than those typical of daytime photography. The compound nature of these distortions renders existing IQA paradigms particularly inadequate.

Score-based IQA methods---from classical metrics like SSIM~\cite{wang2004image} and BRISQUE~\cite{mittal2012no} to recent deep no-reference models~\cite{bosse2017deep,su2020blindly,ke2021musiq,wang2023exploring,zhang2023blind}---condense quality into a single number, which alone cannot distinguish co-occurring distortion types, pinpoint where degradations occur, or explain their physical causes. Multimodal Large Language Models (MLLMs)~\cite{liu2024improved,wang2024qwen2,chen2024internvl,hurst2024gpt} offer a more comprehensive alternative through language-grounded reasoning, and recent works~\cite{wu2024q,wu2024qbench,wu2024qalign,you2024depicting,you2025enhancing} have confirmed their promise for general-purpose quality understanding. However, their effectiveness hinges on domain-specific data---a resource that remains critically absent for nighttime imagery. Mainstream IQA datasets~\cite{hosu2020koniq,fang2020perceptual,lin2019kadid} are overwhelmingly daytime-centric; the few nighttime-oriented resources~\cite{xiang2019blind} provide only coarse Mean Opinion Scores (MOS) without structured annotations; existing language-annotated datasets~\cite{wu2024q,wu2024towards} do not systematically cover nighttime distortions; and current synthetic pipelines apply degradations as uniform post-processing without regard to scene content. We term this joint absence of structured real-world annotations, dedicated evaluation protocols, and scene-aware synthesis for nighttime imagery the \textit{Nighttime Distortion Gap}.

To bridge this gap, we introduce NightIQA, a comprehensive dataset and benchmark suite purpose-built for MLLM-based nighttime IQA. Our contributions are threefold:

\begin{itemize}[noitemsep, leftmargin=*]
    \item \textbf{NightIQA-Real.} A real-world nighttime dataset of 1{,}769 high-resolution images that we \textit{captured ourselves} with modern photographic equipment, spanning three scene categories (\textit{Nature}, \textit{Indoor}, and \textit{Urban}). Six representative distortion types are deliberately induced at three severity levels during shooting, and each image carries structured, multi-dimensional annotations covering distortion type, severity, spatial scope, affected region, and perceptual impact---going well beyond single-score labeling.

    \item \textbf{NightIQA-Bench.} A dedicated evaluation benchmark tailored to MLLM-based nighttime IQA, built around a four-level evaluation hierarchy: (1)~\textit{Perception}---scene parsing at night; (2)~\textit{Technical Assessment}---distortion classification and severity grading; (3)~\textit{Localization}---spatial reasoning about degraded regions; and (4)~\textit{Reasoning}---causal analysis of degradation mechanisms. These four levels probe complementary cognitive dimensions, from scene recognition to causal quality understanding.

    \item \textbf{NightIQA-Syn.} A scene-aware synthesis pipeline that takes high-quality, distortion-free images sourced from Unsplash\cite{unsplash2020dataset} and applies controlled degradations at both global (e.g., sensor noise, underexposure) and instance (e.g., localized motion blur, focus blur) levels, producing spatially coherent artifacts consistent with scene geometry. The pipeline is inherently scalable; we release an initial corpus of 80k high-fidelity images with automatically derived structured annotations alongside the full generation code.
\end{itemize}

To validate NightIQA, we fine-tune Qwen3-VL-8B~\cite{bai2025qwen3} on it via a two-stage pipeline of Supervised Fine-Tuning (SFT) and Group Relative Policy Optimization (GRPO)~\cite{shao2024deepseekmath}. The resulting model substantially outperforms existing open-source IQA methods, matches or exceeds leading closed-source models, and attains the best overall accuracy among all evaluated systems---with a decisive lead on Technical Assessment (L2), precisely the axis where general VLMs falter---indicating that domain-specific data and structured supervision are critical for nighttime IQA.

\section{Related Work}
\label{sec:related}

\begin{figure*}[t]
    \centering
    \includegraphics[width=0.95\linewidth]{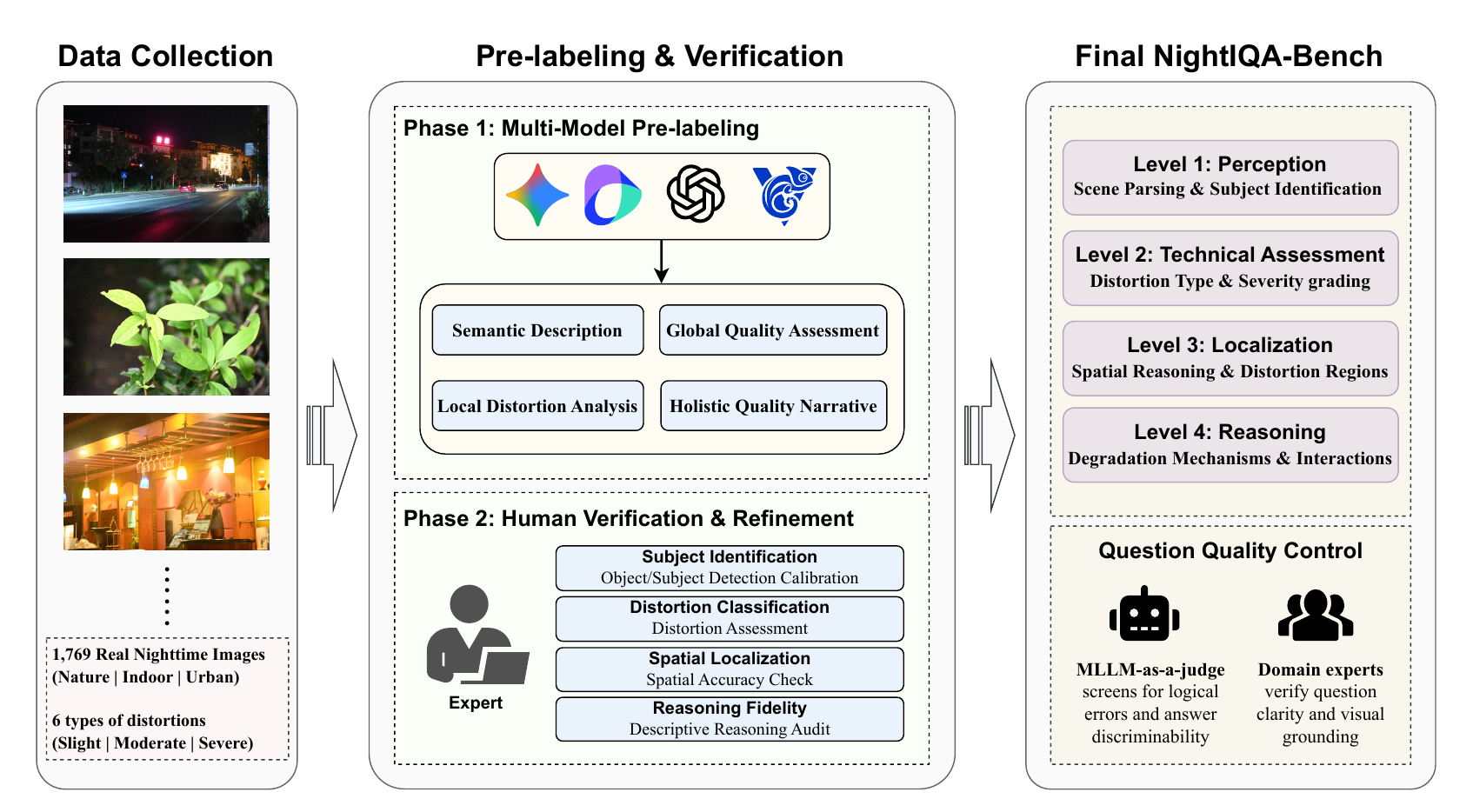}
    \caption{NightIQA-Real annotation and NightIQA-Bench generation in three left-to-right stages. \textit{Data Collection} (left): 1{,}769 real nighttime images across three scene categories and six distortion types. \textit{Pre-labeling \& Verification} (middle): four frontier MLLMs pre-label each image along four content-to-impact dimensions, kept by majority agreement and corrected by experts. \textit{Final NightIQA-Bench} (right): verified labels become a four-level MCQ hierarchy, filtered by an MLLM-as-a-judge with domain-expert validation.}
    \vspace{-4pt}
    \label{fig:annotation_pipeline}
\end{figure*}

\subsection{Image Quality Assessment}
\label{sec:rw_iqa}

IQA methods have evolved along two successive lines.

\noindent\textbf{Score-based approaches} range from full-reference metrics such as SSIM~\cite{wang2004image} and LPIPS~\cite{zhang2018unreasonable} to no-reference estimators like BRISQUE~\cite{mittal2012no} and NIQE~\cite{mittal2012making}. Deep learning further advanced this line through CNN-based score regression~\cite{bosse2017deep,su2020blindly}, multi-scale Transformer architectures~\cite{ke2021musiq,chen2024topiq}, and vision-language cross-modal scoring~\cite{wang2023exploring,zhang2023blind}. While progressively more accurate, all score-based methods inherently condense quality into a single scalar, precluding structured quality understanding.

\noindent\textbf{MLLM-based approaches} reframe IQA as a language generation task, enabling richer output beyond numerical scores. Q-Align~\cite{wu2024qalign} maps images to text-defined quality levels via discrete rating anchors. DeQA-Score~\cite{you2025teaching} models score distributions to capture rating uncertainty. DepictQA~\cite{you2024depicting,you2025enhancing} produces hierarchical descriptions covering severity, causation, and spatial extent. Most recently, Q-Insight~\cite{li2025q} and VisualQuality-R1~\cite{wu2025visualquality} apply Group Relative Policy Optimization (GRPO)~\cite{shao2024deepseekmath} to move beyond the imitation ceiling of supervised fine-tuning, yielding improved reasoning fidelity for quality scoring and distortion perception. Despite this progress, existing MLLM-based IQA methods are developed and evaluated almost exclusively on daytime-centric data; their behavior under compound nighttime degradations remains largely unexamined.

\subsection{IQA Datasets and Benchmarks}
\label{sec:rw_dataset}

\noindent\textbf{Real-world datasets.}
In-the-wild datasets such as KonIQ-10k~\cite{hosu2020koniq} and SPAQ~\cite{fang2020perceptual} collect authentically degraded images from consumer devices and annotate them with scalar MOS labels, offering high ecological validity but limited annotation expressiveness. Language-annotated resources have progressively enriched annotation granularity: Q-Instruct~\cite{wu2024q} provides 200K instruction-response pairs for quality-aware tuning; Q-Bench~\cite{wu2024qbench} and Q-Bench+~\cite{zhang2024q} establish multi-dimensional evaluation covering single-image and paired-comparison settings; Co-Instruct~\cite{wu2024towards} adds comparative quality descriptions; and Q-Ground~\cite{chen2024q} integrates spatial grounding via distortion segmentation. However, these datasets are predominantly collected from daytime scenes, and neither their acquisition strategies nor their annotation schemes target the distortions characteristic of nighttime photography. Even nighttime-specific datasets such as NNID~\cite{xiang2019blind} provide only scalar MOS labels collected with outdated devices, lacking both structured language annotations and modern image fidelity. No existing benchmark systematically evaluates MLLM-based quality understanding under nighttime conditions.

\noindent\textbf{Synthetic datasets.}
Controlled-distortion datasets such as TID2013~\cite{ponomarenko2015image} and KADID-10k~\cite{lin2019kadid} apply predetermined degradations to pristine images, enabling precise control over distortion type and severity. This paradigm has been widely adopted for scalable training data generation. However, existing synthetic pipelines apply degradations as uniform post-processing transforms that ignore scene content, producing artifacts that diverge substantially from real nighttime captures where degradations are shaped by sensor characteristics, scene geometry, and spatially varying illumination.

These observations point to a consistent gap: the community lacks dedicated nighttime data collection, structured annotations, and evaluation protocols on the real-data side, and scene-aware synthesis on the synthetic-data side---the very deficiencies that NightIQA is designed to address.

\section{The NightIQA Dataset}
NightIQA has three coupled components: NightIQA-Real, real nighttime images with multi-dimensional structured annotations; NightIQA-Bench, a benchmark organizing these annotations by cognitive complexity; and NightIQA-Syn, a scalable scene-aware synthetic corpus for training.

\subsection{NightIQA-Real Dataset}
\label{sec:real}

\subsubsection{Data Collection.}
Existing IQA datasets are not merely lacking in nighttime content---they also fail to provide controlled coverage of the distortion types that define nighttime photography. To address both deficiencies, we build NightIQA-Real through controlled real-world capture: 1{,}769 high-resolution nighttime images are shot on location using modern photographic equipment, spanning three scene categories: \textit{Nature}, \textit{Indoor}, and \textit{Urban}. Crucially, rather than passively collecting images of varying quality, we deliberately induce six representative distortion types during capture---\textit{Noise}, \textit{Underexposure}, \textit{Overexposure}, \textit{Motion Blur}, \textit{Focus Blur}, and \textit{Camera Shake}---each at three severity levels (\textit{Slight}, \textit{Moderate}, \textit{Severe}). This controlled acquisition strategy ensures systematic coverage of nighttime degradations while preserving the authenticity of real-world capture conditions. Because the scenes are shot on location, they also retain the complex illumination intrinsic to night photography---specular reflections, and mixed multi-source lighting (e.g., neon, sodium-vapor, and LED sources at differing color temperatures)---conditions that uniform synthetic post-processing cannot faithfully reproduce. \textit{To ensure privacy, all images were strictly screened to exclude clear faces and sensitive details} (Supp.~Sec.~G).

\subsubsection{Annotation Pipeline.}
We label each image along four dimensions that move from scene content to perceptual impact:
(i)~\textit{Semantic Description}, the intended subject and a natural-language account of scene content, layout, and lighting;
(ii)~\textit{Global Quality Assessment}, the primary distortion plus severity ratings for all six canonical types;
(iii)~\textit{Local Distortion Analysis}, each active distortion's spatial scope (\textit{Global} or \textit{Local}) and affected regions;
and (iv)~\textit{Holistic Quality Narrative}, content-impact statements on how active distortions affect perceptual quality and scene understanding. The full annotation schema and field definitions are given in Supp.~Sec.~A.

We annotate in two phases. In Phase~1 (Multi-Model Ensemble Pre-Labeling), four frontier MLLMs from different providers (GPT-5~\cite{singh2025openai}, Gemini 2.5 Pro~\cite{comanici2025gemini}, Doubao-Seed-1.6-Vision~\cite{bytedance2025seed16vision}, and GLM-4.5V~\cite{hong2025glm45v}) annotate each image independently; we keep a label only when at least three of them agree, which provides consistent initial labels for human review. In Phase~2 (Human Verification \& Refinement), professional annotators review every image and correct misidentified distortions, severity ratings, and ungrounded descriptions. This keeps quality close to fully manual annotation at much lower cost; we detail the platform, workflow, and audit protocol in Supp.~Sec.~A.

\subsubsection{Dataset Statistics.}
\begin{figure}[htbp]
  \centering
  \begin{subfigure}{0.49\columnwidth}
    \centering
    \includegraphics[width=\linewidth]{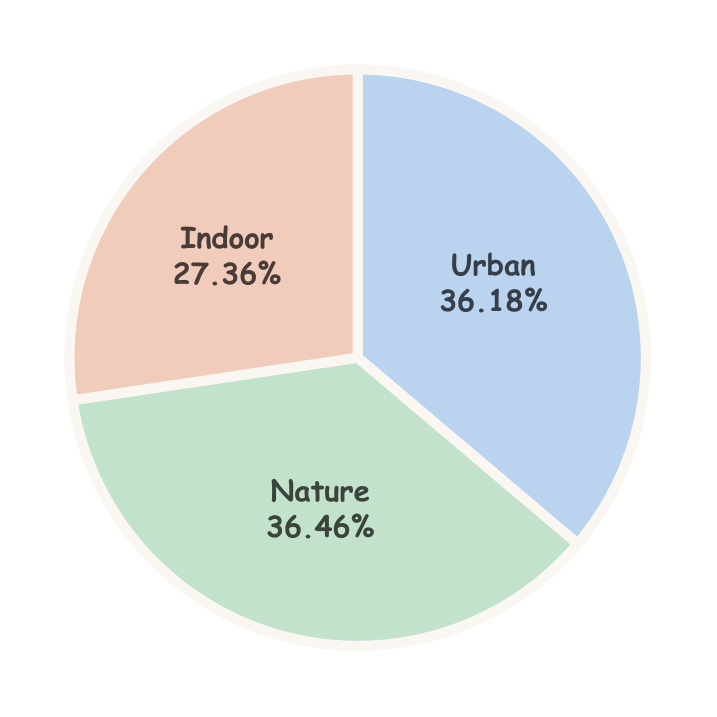}
    \caption{}
    \label{fig:pie_scene}
  \end{subfigure}\hfill
  \begin{subfigure}{0.49\columnwidth}
    \centering
    \includegraphics[width=\linewidth]{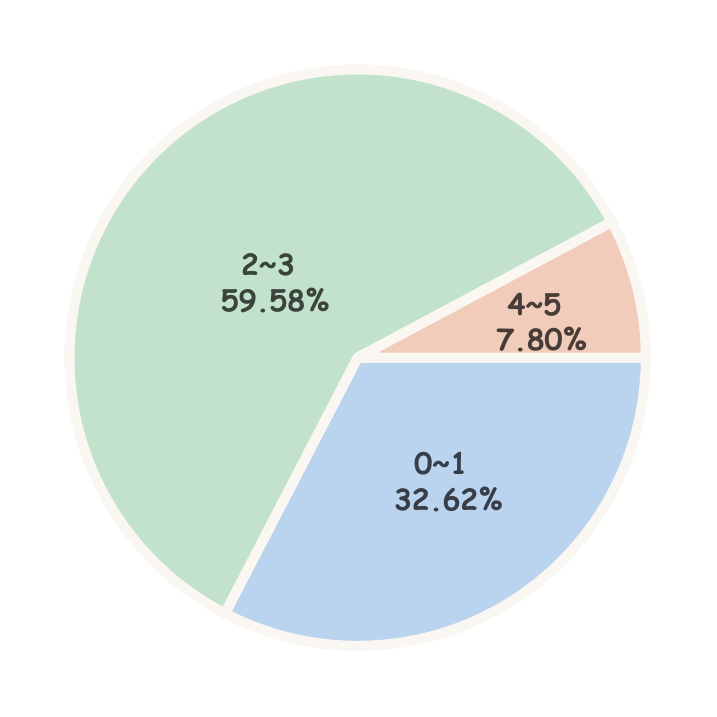}
    \caption{}
    \label{fig:pie_cooccurrence}
  \end{subfigure}
  \caption{NightIQA-Real statistics. (a)~Scene composition: Nature (36.46\%), Urban (36.18\%), Indoor (27.36\%). (b)~Co-occurring distortion types per image: 59.58\% exhibit 2--3 simultaneous distortions.}
  \vspace{-4pt}
  \label{fig:dataset_stats}
\end{figure}

Figure~\ref{fig:dataset_stats} and Table~\ref{tab:distortion_distribution} summarize NightIQA-Real. \textit{Scenes} are balanced across Nature, Urban, and Indoor (Figure~\ref{fig:pie_scene}). \textit{Primary distortions} are mostly exposure problems: Underexposure and Overexposure together exceed 60\% of labels, matching the central difficulty of nighttime capture, while noise and the blurs form a long tail. A small N/A group has no dominant distortion and provides near-pristine references (Table~\ref{tab:distortion_distribution}). \textit{Co-occurrence} is mostly compound: most images carry two or more distortions at once (Figure~\ref{fig:pie_cooccurrence}), which is why we annotate along several dimensions rather than a single distortion label.

\begin{table}[t]
\centering
\caption{Distribution of primary distortion types in NightIQA-Real. N/A denotes images with no visually dominant distortion.}
\vspace{-4pt}
\label{tab:distortion_distribution}
\renewcommand{\arraystretch}{1.0}
\footnotesize
\begin{tabular*}{\columnwidth}{@{\extracolsep{\fill}}lc|c@{}}
\toprule
\textbf{Primary Distortion} & \textbf{Count} & \textbf{Ratio (\%)} \\
\midrule
Underexposure & 767 & 43.36 \\
Overexposure  & 321 & 18.15 \\
Noise         & 180 & 10.18 \\
Camera Shake  & 169 &  9.55 \\
Focus Blur    & 152 &  8.59 \\
N/A           & 105 &  5.94 \\
Motion Blur   &  75 &  4.23 \\
\bottomrule
\end{tabular*}
\vspace{-14pt}
\end{table}

\begin{figure*}[!t]
    \centering
    \includegraphics[width=0.95\linewidth]{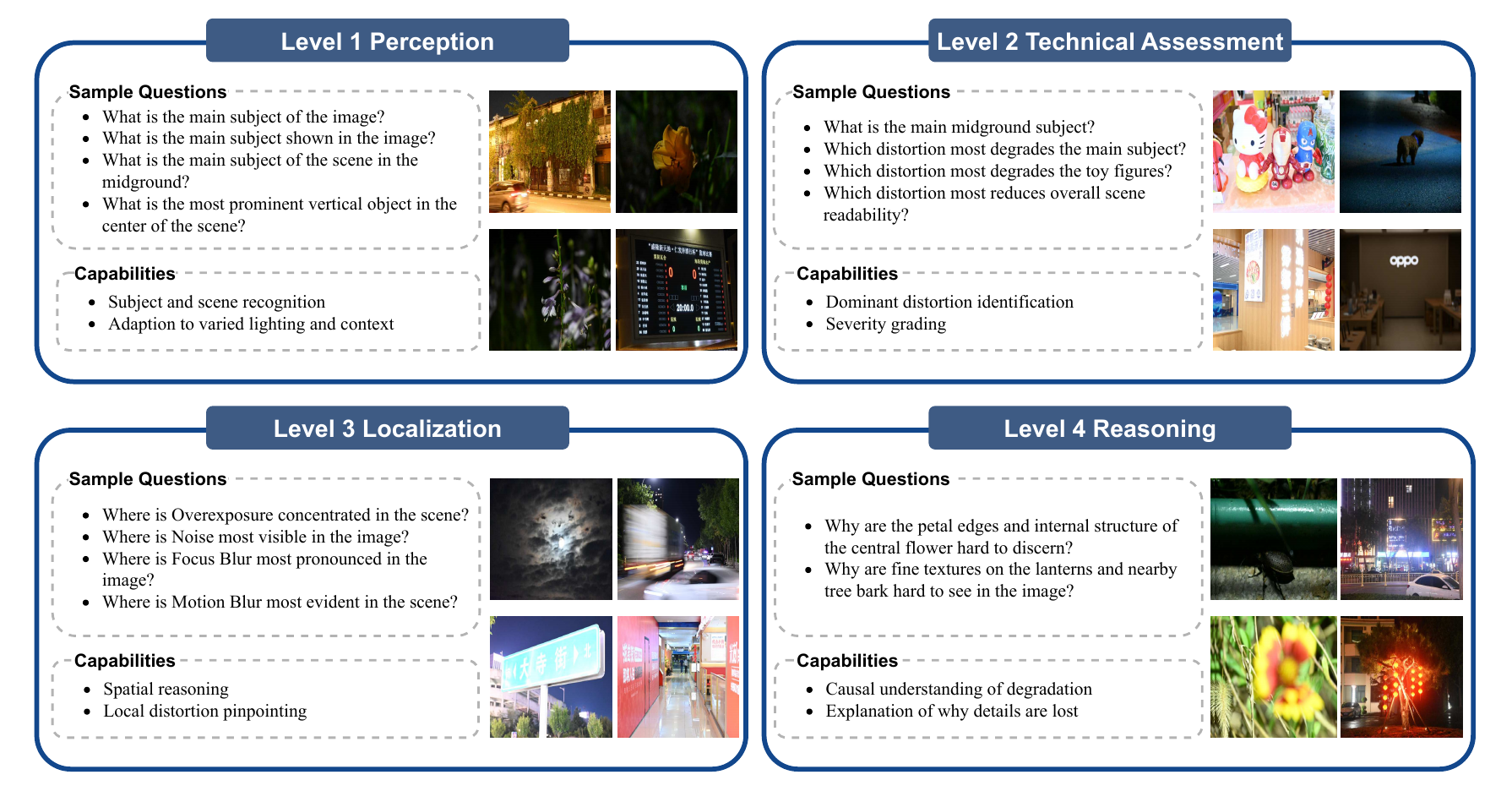}
    \caption{Representative NightIQA-Bench questions across four cognitive levels: L1 (scene recognition), L2 (distortion identification and severity), L3 (spatial reasoning), and L4 (causal degradation analysis).}
    \vspace{-4pt}
    \label{fig:bench}
\end{figure*}

\subsection{NightIQA-Bench: Evaluation Benchmark}
\label{sec:benchmark}

We turn NightIQA-Real's structured, multi-dimensional annotations into targeted evaluation questions to build NightIQA-Bench, a visually grounded multiple-choice question-answering benchmark that tests MLLM capabilities through a four-level evaluation hierarchy (Figure~\ref{fig:annotation_pipeline}).

\subsubsection{Four-Level Evaluation Hierarchy.}
The four levels span complementary cognitive dimensions, mirroring how a human assessor inspects an image, and as Section~\ref{sec:analysis} shows they are not equally difficult: Technical Assessment (L2) is empirically the hardest axis for nearly all models. \textbf{Level 1 (Perception)} recognizes the subject and scene content under nighttime conditions and complex artificial illumination. \textbf{Level 2 (Technical Assessment)} identifies the primary distortion and grades its severity. \textbf{Level 3 (Localization)} locates where a distortion is most evident and ties it to specific image regions. \textbf{Level 4 (Reasoning)} explains why details are degraded, attributing the primary and secondary factors and their interactions.
Representative examples across the four levels are shown in Figure~\ref{fig:bench}. All questions follow a standardized multiple-choice format and are accompanied by a visually grounded rationale derived from observable image evidence. The final validated benchmark comprises 1{,}078 questions, distributed across the four levels as Level 1: 99, Level 2: 344, Level 3: 344, and Level 4: 291. Extended per-level examples and the full filtering pipeline are provided in Supp.~Sec.~C.

\subsubsection{Question Quality Control.}
Benchmark reliability demands that every question is unambiguous, visually answerable, and discriminative. We enforce this through a two-stage filter: an \textit{MLLM-as-a-judge} module first screens candidates for logical errors, weak distractors, and insufficient discriminability; domain experts then manually verify question clarity and visual grounding. Only questions passing both stages are retained.

The training and evaluation splits share no identical or even visually similar scenes. We embed all 1{,}769 images with DINOv2~\cite{oquab2023dinov2} and use feature similarity to assign near-duplicate scenes to the same side of the partition, followed by a manual review that removes any residual visual overlap. This yields a 953-image GRPO training corpus and an 816-image held-out evaluation set with no scene overlap, from which the 1{,}078 benchmark questions are drawn (details in Supp.~Sec.~B).

\subsection{NightIQA-Syn: Scene-Aware Synthetic Data}
\label{sec:synthetic}
\begin{figure*}[!t]
    \centering
    \includegraphics[width=0.95\linewidth]{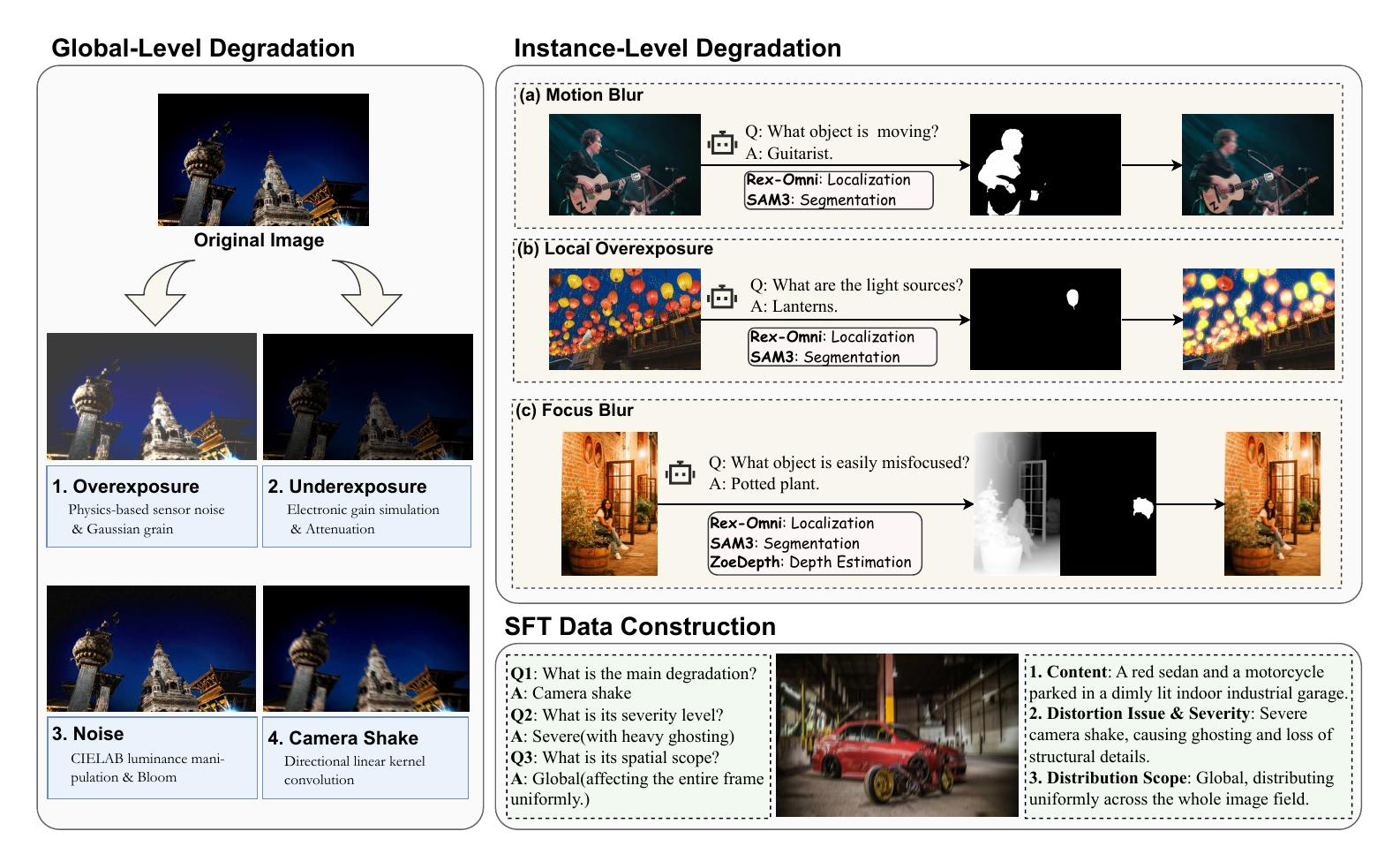}
    \caption{The NightIQA-Syn degradation pipeline. \textit{Left}: global-level distortions are applied uniformly via dedicated operators. \textit{Right}: instance-level distortions follow a three-stage process (target identification, spatial localization, and spatially-coherent rendering) to produce spatially coherent artifacts.}
    \vspace{-4pt}
    \label{fig:syn_pipeline}
\end{figure*}

NightIQA-Real provides high-quality annotations but is limited in scale: real-world collection is costly, scene diversity is bounded, and certain distortion-severity combinations are hard to induce intentionally. NightIQA-Syn complements it with a scene-aware pipeline that generates degraded nighttime images with automatically derived structured labels for large-scale MLLM training. Starting from distortion-free Unsplash~\cite{unsplash2020dataset} sources (clean exposure, sharp focus, no visible artifacts), it applies degradations organized into two categories by their spatial extent. \textit{Global-level} distortions come from sensor or camera behavior and affect the whole image uniformly, whereas \textit{instance-level} distortions are tied to specific scene elements and follow scene geometry. Each distortion is produced by a dedicated operator (Figure~\ref{fig:syn_pipeline}).

\begin{table*}[!t]
\centering
\caption{Performance comparison on NightIQA-Bench (MCQ accuracy, \%). L1--L4: Perception, Technical Assessment, Localization, and Reasoning. Best in \textbf{bold}, second best \underline{underlined}.}
\vspace{-4pt}
\label{tab:main_results}
\renewcommand{\arraystretch}{1.17}
\setlength{\tabcolsep}{12pt}
\resizebox{\textwidth}{!}{%
\begin{tabular}{llcccc|c}
\toprule
\textbf{Type} & \textbf{Model} & \textbf{L1: Percep.} & \textbf{L2: Tech.} & \textbf{L3: Local.} & \textbf{L4: Reason.} & \textbf{Avg.} \\
\midrule
\multicolumn{2}{l}{\textbf{Sampled random guess}} & 25.89 & 25.09 & 25.19 & 24.88 & 25.26 \\
\midrule
\multirow{6}{*}{\textbf{Closed-Source}}
  & GPT-4o~\cite{hurst2024gpt}                & 81.82 & 71.22 & 71.51 & 75.95 & 73.56 \\
  & GPT-5 Mini~\cite{singh2025openai}            & 83.84 & 66.57 & 78.20 & 75.26 & 74.21 \\
  & Gemini-2.5-Flash~\cite{comanici2025gemini}      & 71.72 & 48.26 & 61.92 & 61.86 & 58.44 \\
  & Gemini-3.1-Flash-Lite~\cite{google2025gemini31} & \textbf{89.90} & 73.26 & 81.10 & 79.38 & 78.94 \\
  & Doubao-Seed-2.0-Mini~\cite{bytedance2026seed20mini}  & 86.87 & 62.50 & 75.29 & 70.79 & 71.06 \\
  & Claude-Haiku-4.5~\cite{anthropic2025claudehaiku}      & 72.73 & 64.24 & 68.90 & 74.23 & 69.20 \\
\midrule
\multirow{12}{*}{\textbf{Open-Source}}
  & MiMo-V2-Omni~\cite{xiaomi2026mimo}          & 83.84 & 69.77 & 73.84 & 73.88 & 73.47 \\
  & MiMo-V2.5~\cite{xiaomi2026mimov25}             & 77.78 & 68.02 & 75.29 & 74.57 & 73.01 \\
  & GLM-4.5V~\cite{hong2025glm45v}              & 84.85 & 61.63 & 76.16 & 69.76 & 70.59 \\
  & GLM-4.6V~\cite{zai2025glm46v}              & 82.83 & 63.95 & 74.71 & 68.04 & 70.22 \\
  & Qwen3-VL-30B~\cite{bai2025qwen3}          & 82.83 & 59.59 & 70.93 & 67.70 & 67.53 \\
  & Qwen3-VL-235B~\cite{bai2025qwen3}         & 86.87 & 65.41 & 80.52 & 77.66 & 75.51 \\
  & Qwen3.5-27B~\cite{team2026qwen3}           & 86.87 & 69.77 & 81.69 & 74.91 & 76.53 \\
  & Qwen3.5-35B-A3B~\cite{team2026qwen3}       & \underline{88.89} & 69.19 & \textbf{82.85} & \underline{80.76} & 78.48 \\
  & Qwen3.5-122B-A10B~\cite{team2026qwen3}     & 83.84 & 66.86 & 79.94 & 75.26 & 74.86 \\
  & Qwen3.5-397B~\cite{team2026qwen3}          & 86.87 & \underline{74.42} & \underline{82.27} & \textbf{82.47} & \underline{80.24} \\
  & InternVL3.5-8B~\cite{wang2025internvl35}        & 70.71 & 52.33 & 57.85 & 63.57 & 58.81 \\
  & InternVL3.5-38B~\cite{wang2025internvl35}       & 84.85 & 61.92 & 70.35 & 65.98 & 67.81 \\
\midrule
\multirow{4}{*}{\textbf{IQA}}
  & DepictQA~\cite{you2024depicting}              & 57.60 & 42.70 & 41.60 & 62.50 & 49.10 \\
  & Q-Instruct~\cite{wu2024q}            & 46.46 & 45.93 & 53.78 & 56.70 & 51.39 \\
  & VisualQuality-R1~\cite{wu2025visualquality}      & 66.70 & 55.20 & 63.70 & 59.50 & 61.30 \\
  & Q-Insight~\cite{li2025q}             & 65.70 & 51.20 & 59.60 & 62.50 & 59.70 \\
\midrule
\multirow{1}{*}{\textbf{Ours}}
  & SFT + GRPO (8B)       & 87.88 & \textbf{78.20} & 81.69 & 80.07 & \textbf{80.71} \\
\bottomrule
\end{tabular}%
}
\end{table*}

\subsubsection{Global-Level Degradation.}
We model four types, partly via Albumentations~\cite{buslaev2020albumentations} (Figure~\ref{fig:syn_pipeline}, left): \textit{Noise} (chromatic and Gaussian sensor noise typical of nighttime capture); \textit{Underexposure} (intensity attenuation with added noise); \textit{Global Overexposure} (luminance boosting with highlight bloom); and \textit{Camera Shake} (directional motion blur with randomized extent and orientation). Exact operators, sampling ranges, and per-severity parameters are detailed in Supp.~Sec.~D.1.

\subsubsection{Instance-Level Degradation.}
Some nighttime distortions are inherently localized: motion blur affects only moving objects, focus blur varies with depth, and overexposure blooms around point light sources. Applying them uniformly would misrepresent how they appear in real captures. Our instance-level pipeline (Figure~\ref{fig:syn_pipeline}, right) proceeds in three stages.
\textit{(1)~Target identification}: a vision-language model checks each image for elements suitable to a given distortion (e.g., movable objects for motion blur, likely-misfocused objects for focus blur, bright point sources for local overexposure) and names the candidate targets it finds; only images with valid targets enter the corresponding branch.
\textit{(2)~Spatial localization}: Rex-Omni~\cite{jiang2025detect} grounds these candidate targets via open-vocabulary localization and SAM3~\cite{carion2025sam} extracts pixel-accurate masks, with ZoeDepth~\cite{bhat2023zoedepth} supplying monocular depth for depth-dependent effects such as focus blur.
\textit{(3)~Spatially-coherent rendering}: dedicated operators synthesize spatially varying degradations. \textit{Local Overexposure} adds atmospheric bloom and glare halos around bright sources, \textit{Motion Blur} accumulates warped samples along randomized trajectories in linear color space, and \textit{Focus Blur} applies depth-aware, spatially varying blur radii for perceptually plausible bokeh. Full algorithms and parameter ranges are provided in Supp.~Sec.~D.2.

\subsubsection{SFT Data Construction.}
We release an initial 80{,}000-image corpus with full generation code. Each image carries automatically generated structured annotations in two formats: \textit{Basic QA}, three single-turn pairs probing distortion type, severity, and spatial scope; and \textit{Chain-of-Thought (CoT)}, a long-form reasoning sample that describes image content, identifies the distortion, and characterizes its spatial distribution. We sample 6{,}769 images for SFT, balancing distortion-type and severity coverage (Supp.~Sec.~D.3), stored as conversation-style records compatible with standard MLLM fine-tuning frameworks.

\section{Baseline Construction}
\label{sec:method}

We build a baseline on Qwen3-VL-8B~\cite{bai2025qwen3} trained on NightIQA in two stages: SFT on the synthetic NightIQA-Syn data, followed by GRPO on the real NightIQA-Real annotations. We keep the backbone architecture unchanged and intend the pipeline as a reproducible reference rather than a new model.

\subsection{Stage 1: Supervised Fine-Tuning (SFT)}
\label{sec:sft}

We first perform SFT on the structured instruction data from NightIQA-Syn, enabling the model to learn nighttime distortion types, severity levels, and spatial scopes from scalable synthetic supervision. We use the standard masked next-token prediction loss:
\begin{equation}
  \mathcal{L}_{\text{SFT}} = -\sum_{t=1}^{T} m_t \log \pi_{\theta}(z_t \mid z_{<t}, \tilde{x}, q),
  \label{eq:sft_loss}
\end{equation}
where $\mathbf{z} = (z_1, \dots, z_T)$ is the target response, $m_t$ masks non-response tokens, and $\pi_{\theta}$ denotes the model policy. The SFT data construction, training objective, and full hyperparameters are detailed in Supp.~Sec.~E.

\subsection{Stage 2: GRPO-Based Alignment}
\label{sec:grpo}

We then apply GRPO~\cite{shao2024deepseekmath} using NightIQA-Real to align the model with human-annotated reasoning traces. For each image--question pair, the model samples multiple candidate responses, which are scored by a task-specific composite reward. The complete GRPO objective, advantage estimation, training configuration, and system prompt are deferred to Supp.~Sec.~F.

\noindent\textbf{Composite Reward Function.}
The reward encourages both correct answers and faithful reasoning. We use a structured CoT format and define $r_i$ as a weighted sum of answer correctness, reasoning fidelity, and format compliance:
\begin{equation}
  r_i = w_{\text{acc}} r_{\text{acc}} + w_{\text{desc}} r_{\text{desc}} + w_{\text{fmt}} r_{\text{fmt}},
\end{equation}
where $w_{\mathrm{acc}} = 1.0$, $w_{\mathrm{desc}} = 0.8$, and $w_{\mathrm{fmt}} = 0.5$. Here, $r_{\mathrm{acc}}$ checks answer correctness, $r_{\mathrm{desc}}$ measures reasoning similarity to the human annotations using BLEU-4~\cite{papineni2002bleu} and ROUGE-L~\cite{lin2004rouge}, and $r_{\mathrm{fmt}}$ enforces the required output format.

\section{Benchmark Evaluation}
\label{sec:evaluation}

\subsection{Evaluation Setup}
\label{sec:eval_setup}
We benchmark 22 zero-shot baselines across three families: \textit{closed-source frontier models} (GPT-4o~\cite{hurst2024gpt}, GPT-5 Mini~\cite{singh2025openai}, Gemini~\cite{comanici2025gemini, google2025gemini31}, Doubao~\cite{bytedance2026seed20mini}, Claude~\cite{anthropic2025claudehaiku}), \textit{open-source VLMs} (MiMo~\cite{xiaomi2026mimo, xiaomi2026mimov25}, GLM-4.5V/4.6V~\cite{hong2025glm45v, zai2025glm46v}, Qwen3-VL~\cite{bai2025qwen3}, Qwen3.5~\cite{team2026qwen3}, InternVL3.5~\cite{wang2025internvl35}), and \textit{domain-specific IQA models} (DepictQA~\cite{you2024depicting}, Q-Instruct~\cite{wu2024q}, VisualQuality-R1~\cite{wu2025visualquality}, Q-Insight~\cite{li2025q}); see Table~\ref{tab:main_results} for exact variants.
We also report final SFT+GRPO in Table~\ref{tab:main_results} and analyze GRPO-only in the ablation. We report MCQ accuracy on L1--L4 and overall accuracy over all 1{,}078 held-out questions from the 816-image scene-disjoint evaluation set.
\subsection{Results and Analysis}
\label{sec:analysis}

Across all evaluated models, NightIQA-Bench surfaces four consistent patterns (Table~\ref{tab:main_results}, Figure~\ref{fig:benchmark_petal}).

\noindent\textbf{Cognitive Stratification Validates Benchmark Design.}
Accuracy is far from uniform: most general VLMs peak on Perception (L1) and bottom out on Technical Assessment (L2). Even the top Perception scorer, Gemini-3.1-Flash-Lite~\cite{google2025gemini31}, slides from 89.90\% (L1) to 73.26\% (L2). General VLMs parse nighttime scenes but lack the expertise to classify their distortions.

\noindent\textbf{Domain-Specific IQA Models Exhibit a Severe Gap.}
Dedicated IQA methods cluster at the bottom: DepictQA~\cite{you2024depicting} (49.10\%), Q-Instruct~\cite{wu2024q} (51.39\%), Q-Insight~\cite{li2025q} (59.70\%), and VisualQuality-R1~\cite{wu2025visualquality} (61.30\%) trail nearly all general VLMs and falter on quality-centric axes such as Technical Assessment (L2) and Localization (L3). Their daylight-centric training with generic synthetic degradations (e.g., Gaussian blur, JPEG) leaves them ill-equipped for nighttime artifacts such as color noise and uneven exposure.

\noindent\textbf{Scaling Alone Cannot Close the Nighttime Gap.}
Parameter scaling yields inconsistent, saturating returns: within Qwen3.5 accuracy is non-monotonic in size, and the largest Qwen3.5-397B~\cite{team2026qwen3} (80.24\%) is still edged out by our domain-aligned 8B model, with its shortfall sharpest on L2 (74.42\%). These results suggest that scaling alone is insufficient, and domain-specific degradation knowledge is a critical missing factor.

\noindent\textbf{NightIQA Brings an 8B Model to Frontier-Level Accuracy.}
Our SFT+GRPO model attains the best overall average (80.71\%), surpassing the ${\sim}50\times$ larger Qwen3.5-397B (80.24\%) and the strongest closed-source model, Gemini-3.1-Flash-Lite (78.94\%), with a clear lead on Technical Assessment (L2, 78.20\%) where general VLMs drop sharply. Larger general VLMs keep narrow leads on single axes (Gemini-3.1-Flash-Lite on L1, Qwen3.5-35B-A3B on L3, Qwen3.5-397B on L4), but none matches our overall accuracy and L2 lead, supporting our data-centric motivation.

\subsection{Ablation Study}
\label{sec:ablation}

We ablate each training stage and data source from the same Qwen3-VL-8B base model (Table~\ref{tab:ablation}).

\begin{table}[htbp]
\centering
\small
\caption{Ablation on the training pipeline. SFT uses NightIQA-Syn; GRPO uses NightIQA-Real. $\Delta$: improvement over the preceding row.}
\vspace{-4pt}
\label{tab:ablation}
\renewcommand{\arraystretch}{1.0}
\resizebox{\columnwidth}{!}{%
\begin{tabular}{lcccc|c}
\toprule
\textbf{Configuration} & \textbf{L1} & \textbf{L2} & \textbf{L3} & \textbf{L4} & \textbf{Avg.} \\
\midrule
Qwen3-VL-8B (base)     & 71.72 & 59.30 & 64.83 & 65.29 & 63.82 \\
\midrule
+ GRPO (Real)           & 82.83 & 75.00 & 79.65 & 74.91 & 77.18 \\
\quad $\Delta$          & {\small\textcolor{teal}{+11.1}} & {\small\textcolor{teal}{+15.7}} & {\small\textcolor{teal}{+14.8}} & {\small\textcolor{teal}{+9.6}} & {\small\textcolor{teal}{+13.4}} \\
\midrule
+ SFT (Syn) + GRPO (Real)     & \textbf{87.88} & \textbf{78.20} & \textbf{81.69} & \textbf{80.07} & \textbf{80.71} \\
\quad $\Delta$          & {\small\textcolor{teal}{+5.1}} & {\small\textcolor{teal}{+3.2}} & {\small\textcolor{teal}{+2.0}} & {\small\textcolor{teal}{+5.2}} & {\small\textcolor{teal}{+3.5}} \\
\bottomrule
\end{tabular}%
}
\vspace{-2pt}
\end{table}

\noindent\textbf{Effect of GRPO (NightIQA-Real).}
GRPO alone adds 13.4\,pp to the average (63.82\% to 77.18\%), with the largest gains on L2 (+15.7) and L3 (+14.8): NightIQA-Real's reasoning trajectories are the dominant driver in our ablation, teaching distortion recognition and spatial reasoning at night.

\noindent\textbf{Effect of SFT (NightIQA-Syn).}
Prepending SFT lifts the average a further 3.5\,pp (77.18\% to 80.71\%), concentrated on L4 (+5.2) and L1 (+5.1). These gains stack on top of GRPO's: NightIQA-Syn's large-scale distortion vocabulary adds supervision beyond NightIQA-Real's reasoning traces.

\noindent\textbf{Training dynamics across epochs.}
Both regimes generally improve before saturating or slightly regressing (Table~\ref{tab:epoch}). SFT+GRPO climbs from 78.57\% (epoch~1) to a peak 80.71\% at epoch~4, which is also its best L4 (80.07\%), then dips to 80.43\% at epoch~5; GRPO-only peaks later (77.18\%, epoch~5). We therefore adopt SFT+GRPO at epoch~4 for its best average and best L4 (Reasoning), the highest-order and most challenging axis.

\begin{table}[t]
\centering
\small
\caption{Per-epoch accuracy (\%) for GRPO-only (from base) and SFT+GRPO. Best per column in \textbf{bold}: SFT+GRPO leads on every level. We adopt it at epoch~4 for its best average and L4 (Reasoning); GRPO-only peaks at epoch~5.}
\vspace{-4pt}
\label{tab:epoch}
\renewcommand{\arraystretch}{1.0}
\resizebox{\columnwidth}{!}{%
\begin{tabular}{clcccc|c}
\toprule
\textbf{Method} & \textbf{Epoch} & \textbf{L1} & \textbf{L2} & \textbf{L3} & \textbf{L4} & \textbf{Avg.} \\
\midrule
\multirow{5}{*}{GRPO-only}
  & 1 & 79.80 & 64.53 & 75.00 & 71.13 & 71.06 \\
  & 2 & 78.79 & 70.93 & 75.58 & 76.29 & 74.58 \\
  & 3 & 85.86 & 70.06 & 77.91 & 76.29 & 75.70 \\
  & 4 & 82.83 & 70.06 & 77.91 & 77.32 & 75.70 \\
  & 5 & 82.83 & 75.00 & 79.65 & 74.91 & 77.18 \\
\midrule
\multirow{5}{*}{SFT+GRPO}
  & 1 & 80.81 & \textbf{78.20} & 78.49 & 78.35 & 78.57 \\
  & 2 & 80.81 & 76.74 & 80.81 & 79.04 & 79.04 \\
  & 3 & \textbf{88.89} & 75.00 & \textbf{82.27} & 79.38 & 79.78 \\
  & 4 & 87.88 & \textbf{78.20} & 81.69 & \textbf{80.07} & \textbf{80.71} \\
  & 5 & \textbf{88.89} & 76.74 & \textbf{82.27} & 79.73 & 80.43 \\
\bottomrule
\end{tabular}%
}
\vspace{-6pt}
\end{table}

\section{Conclusion}
\label{sec:conclusion}

We present NightIQA: 1,769 originally captured, multi-dimensionally annotated nighttime images; a four-level benchmark spanning perception to causal reasoning; and a scene-aware synthesis pipeline releasing 80k images with code. Across 22 zero-shot baselines, we expose a severe nighttime domain gap that scaling alone does not close. Fine-tuning Qwen3-VL-8B with SFT and GRPO yields a 17-point gain and the best overall accuracy of any evaluated system, including models ${\sim}50\times$ larger, supporting the importance of domain-specific data and structured supervision. A natural next step is to grow NightIQA beyond the nighttime regime, extending the dataset to daytime and broader real-world scenarios and enriching it with additional scenes, distortion types, and annotation dimensions, so that it can serve as a foundation for general MLLM-based image quality assessment.

{
    \small
    \fontencoding{T1}\selectfont
    \urlstyle{same}

}
\clearpage
\newcommand{\suppref}[1]{Supp.~Sec.~\ref{#1}}
\setcounter{figure}{5}
\setcounter{table}{4}
\setcounter{equation}{2}

\appendix

\newsavebox{\nightjsonbox}
\newsavebox{\nightjsonexbox}
\newsavebox{\codeboxglobal}
\newsavebox{\codeboxmotion}
\newsavebox{\codeboxfocus}
\newsavebox{\codeboxlocal}
\newsavebox{\nighttargetbox}
\newsavebox{\nightvalbox}
\newcommand{\ptag}[1]{{\small\bfseries\textsc{\MakeLowercase{#1}}.}}
\newcommand{\cardtitle}[1]{\rule{\textwidth}{0.9pt}\\[2.5pt]{\small\bfseries #1}\\[2.5pt]\rule{\textwidth}{0.4pt}}
\newcommand{\botrule}{\rule{\textwidth}{0.9pt}}
\newcommand{\codecard}[1]{\usebox{#1}}
\newcommand{\prosecard}[1]{\begin{minipage}{\textwidth}#1\end{minipage}}
\definecolor{benchok}{rgb}{0.0,0.42,0.0}
\newcommand{\correct}{~\textcolor{benchok}{\textbf{[correct]}}}
\newcommand{\benchpanel}[3]{%
  \rule{\textwidth}{0.9pt}\\[2pt]%
  {\small\bfseries #1}\\[2pt]%
  \rule{\textwidth}{0.4pt}\\[5pt]%
  \begin{minipage}[c]{0.31\textwidth}\centering\includegraphics[width=\textwidth]{#2}\end{minipage}%
  \hfill
  \begin{minipage}[c]{0.655\textwidth}\footnotesize #3\end{minipage}\\[4pt]%
  \rule{\textwidth}{0.9pt}}
\newcommand{\bquestion}[1]{\textbf{Question.}~#1\\[2pt]}
\newcommand{\bcot}[1]{\textbf{Chain-of-Thought.}~#1}
\newenvironment{boptions}{\begin{enumerate}[label=(\Alph*),leftmargin=1.7em,itemsep=1pt,topsep=2pt,parsep=0pt]}{\end{enumerate}\vspace{2pt}}

\renewcommand{\topfraction}{0.92}
\renewcommand{\bottomfraction}{0.85}
\renewcommand{\textfraction}{0.08}
\renewcommand{\floatpagefraction}{0.72}
\renewcommand{\dbltopfraction}{0.92}
\renewcommand{\dblfloatpagefraction}{0.72}
\setcounter{topnumber}{3}
\setcounter{bottomnumber}{2}
\setcounter{totalnumber}{5}
\setcounter{dbltopnumber}{3}

\begin{center}
    {\Large\bfseries Supplementary Material}\\[0.35em]
    {\normalsize\itshape NightIQA: A Comprehensive Dataset and Benchmark\\ for MLLM-Based Nighttime Image Quality Assessment}
\end{center}
\noindent\rule{\columnwidth}{0.6pt}
\vspace{0.6em}

\noindent This supplement collects the implementation details, full derivations, and extended results deferred from the main paper. It is organized as follows:

\begin{itemize}[noitemsep, leftmargin=1.4em]
    \item \textbf{Sec.~\ref{app:annotation}}~--~NightIQA-Real annotation platform, workflow, and quality audit.
    \item \textbf{Sec.~\ref{app:dedup}}~--~scene-level deduplication and leakage-free train/benchmark partitioning.
    \item \textbf{Sec.~\ref{app:bench}}~--~NightIQA-Bench question statistics, filtering pipeline, and per-level examples.
    \item \textbf{Sec.~\ref{app:syn}}~--~\textbf{full NightIQA-Syn degradation algorithms, parameters, and core reference implementation} (global- and instance-level).
    \item \textbf{Sec.~\ref{app:sft}}~--~supervised fine-tuning data, objective, and hyperparameters.
    \item \textbf{Sec.~\ref{app:grpo}}~--~complete GRPO formulation, composite reward, and training configuration.
    \item \textbf{Sec.~\ref{app:ethics}}~--~ethical considerations, privacy protection, and licensing.
\end{itemize}
\vspace{0.4em}

\section{NightIQA-Real Annotation Details}
\label{app:annotation}

\subsection{Annotation Schema and Field Definitions}
\label{app:annotation_schema}

Each NightIQA-Real image is labeled along the four structured dimensions introduced in Section~\ref{sec:real}, progressing from scene content to perceptual impact. Table~\ref{tab:annotation_schema} enumerates every field together with its admissible values.

\begin{enumerate}[noitemsep, leftmargin=1.4em]
    \item \textbf{Semantic Description} --- the intended subject together with a natural-language account of scene content, spatial layout, and lighting conditions.
    \item \textbf{Global Quality Assessment} --- the single primary distortion of the image, together with severity ratings for the six canonical distortion types.
    \item \textbf{Local Distortion Analysis} --- for each active distortion, its spatial scope (\textit{Global} or \textit{Local}) and the affected image region(s).
    \item \textbf{Holistic Quality Narrative} --- content-impact statements describing how the active distortions affect perceptual quality and scene understanding.
\end{enumerate}

\begin{table}[htbp]
    \centering
    \caption{NightIQA-Real annotation schema. Distortion types, severity levels, spatial scopes, and scene categories follow the definitions used during capture (Section~\ref{sec:real}).}
    \label{tab:annotation_schema}
    \footnotesize
    \begin{tabular}{@{}p{0.26\columnwidth} p{0.30\columnwidth} p{0.30\columnwidth}@{}}
        \toprule
        \textbf{Dimension} & \textbf{Field} & \textbf{Values} \\
        \midrule
        \multirow{3}{*}{Semantic Desc.}
            & Subject                      & free text \\
            & Scene category               & Nature / Indoor / Urban \\
            & Content, layout \& lighting  & free text \\
        \midrule
        \multirow{2}{*}{Global Quality}
            & Primary distortion           & one of the six types, or N/A \\
            & Per-type severity            & Slight / Moderate / Severe \\
        \midrule
        \multirow{2}{*}{Local Distortion}
            & Spatial scope                & Global / Local \\
            & Affected region(s)           & free text \\
        \midrule
        Holistic Narrative
            & Content-impact statement     & free text \\
        \bottomrule
    \end{tabular}
\end{table}

The six canonical distortion types are \textit{Noise}, \textit{Underexposure}, \textit{Overexposure}, \textit{Motion Blur}, \textit{Focus Blur}, and \textit{Camera Shake}, each induced at three severity levels (\textit{Slight}, \textit{Moderate}, \textit{Severe}) during capture; the N/A label denotes an image with no visually dominant distortion (Table~\ref{tab:distortion_distribution}).

\subsection{Annotation Platform and Workflow}
\label{app:annotation_workflow}

We employ \textbf{Label Studio}\footnote{\url{https://labelstud.io/}} as the annotation platform, implementing the two-phase protocol of Section~\ref{sec:real}: \textbf{Phase~1 (Multi-Model Ensemble Pre-Labeling)} supplies the imported pre-fills in step~(1), and \textbf{Phase~2 (Human Verification \& Refinement)} comprises steps~(2)--(4) below.

\begin{enumerate}[noitemsep, leftmargin=1.6em, label=\textbf{(\arabic*)}]
    \item \textbf{Pre-labeling Import.} Ensemble annotations from the four frontier MLLMs (GPT-5, Gemini~2.5~Pro, Doubao-Seed-1.6-Vision, and GLM-4.5V), retained only where at least three models agree (Section~\ref{sec:real}), are imported as pre-filled fields.
    \item \textbf{Human Verification.} Annotators review each image alongside pre-filled dropdowns, radio buttons, and free-text fields.
    \item \textbf{Correction.} Misidentified distortions, severity ratings, and ungrounded descriptions are corrected; disagreements are escalated to a senior annotator.
    \item \textbf{Quality Audit.} Finalized annotations are exported in JSON, and a random 10\% subset undergoes a second-round audit.
\end{enumerate}

\begin{figure*}[tp]
    \centering
    \begin{subfigure}[t]{\textwidth}
        \centering
        \includegraphics[width=\textwidth]{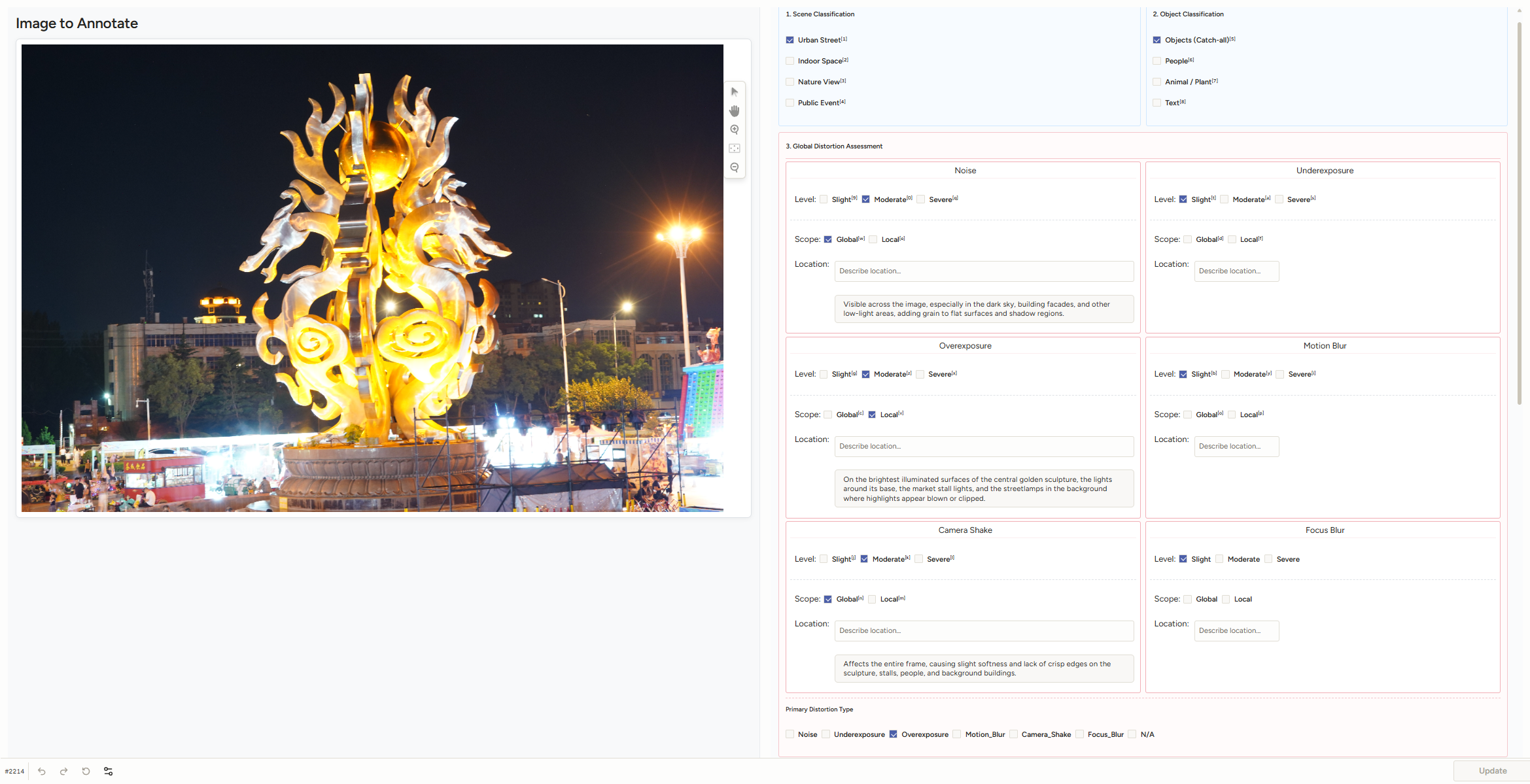}
    \end{subfigure}
    \vspace{0.3em}
    \begin{subfigure}[t]{\textwidth}
        \centering
        \includegraphics[width=\textwidth]{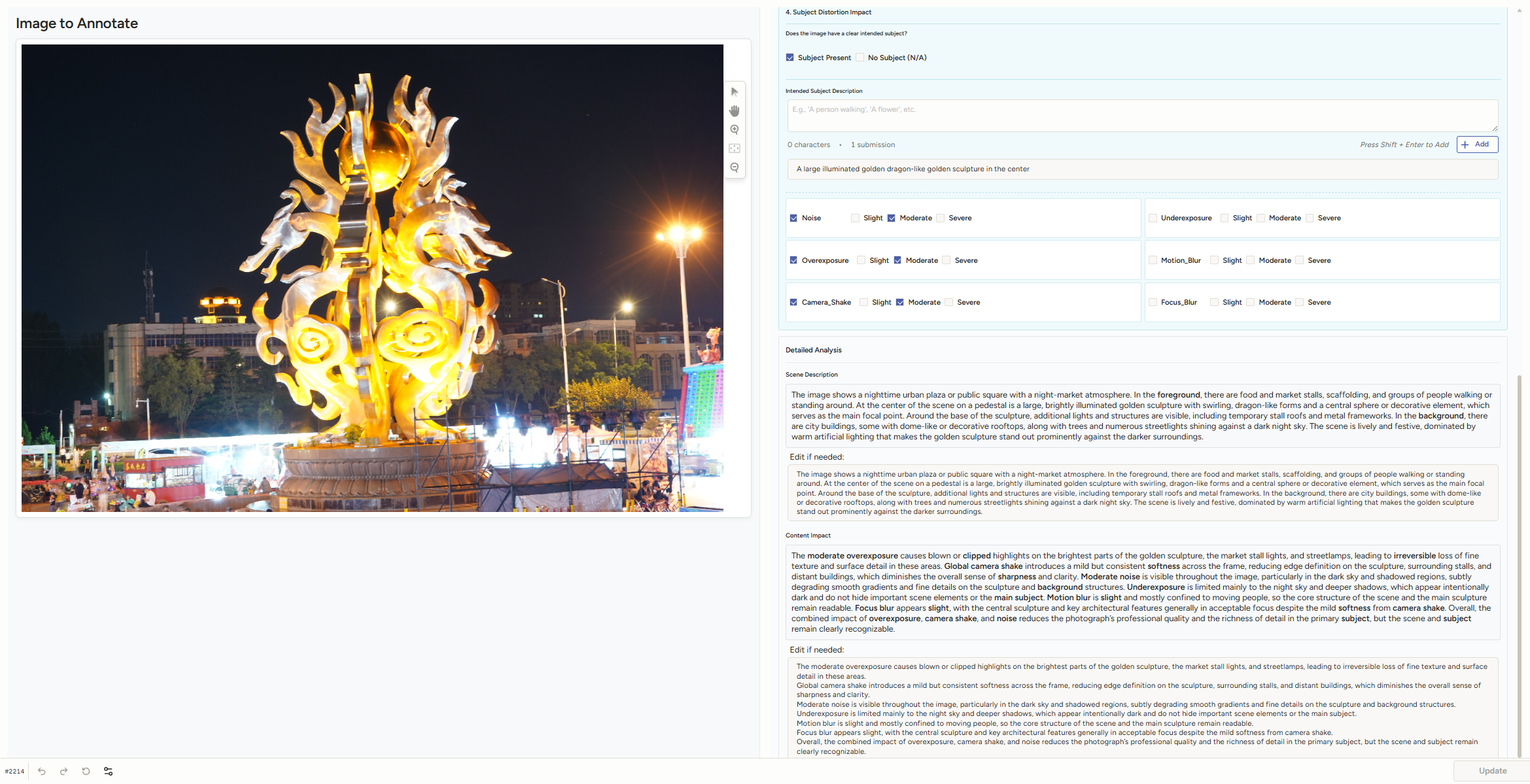}
    \end{subfigure}
    \caption{Screenshots of the Label Studio annotation interface used for NightIQA-Real. The interface presents each nighttime image alongside pre-filled annotations from the multi-model ensemble, enabling annotators to verify and refine distortion type, severity, spatial scope, and quality narratives.}
    \label{fig:label_studio}
\end{figure*}

\noindent Figure~\ref{fig:label_studio} shows representative screenshots of the interface.

\subsection{Pre-Labeling Prompt Template and Annotation Schema}
\label{app:annotation_prompt}

Phase~1 of the annotation pipeline (Section~\ref{sec:real}) queries the four frontier MLLMs (GPT-5, Gemini~2.5~Pro, Doubao-Seed-1.6-Vision, GLM-4.5V) with the \emph{identical} structured prompt of Figure~\ref{fig:annotation_prompt}; each model returns a structured JSON annotation following the same schema, and a field is retained only where at least three models agree. The prompt fixes the task definition, a priority rule for selecting the primary distortion, the severity vocabulary, and the exact output schema, so that the four models produce directly comparable structured outputs. Figure~\ref{fig:annotation_example} shows a complete annotation produced by this process and verified by human annotators.

\begin{lrbox}{\nightjsonbox}
\begin{minipage}{\textwidth}
\ttfamily\footnotesize
\begin{verbatim}
{
  "scene_category": "few words, e.g. Outdoor street view",
  "distortion_severity": {
      "Noise": "Severe|Moderate|Slight",
      "Underexposure": "...", "Overexposure": "...",
      "Motion_Blur": "...",   "Camera_Shake": "...",
      "Focus_Blur": "..."
  },
  "primary_distortion": "single dominant Severe/Moderate type",
  "intended_subject": "single main focal subject, or N/A",
  "subject_distortions": { "type": "severity", ... },
  "scene_description": "key elements + spatial positions",
  "distortion_analysis": {
      "type": "Global|Local - short location phrase", ...
  },
  "content_impact_discussion": "impact on subject + scene",
  "explanation": ["concise justification of key choices", ...]
}
\end{verbatim}
\end{minipage}
\end{lrbox}

\begin{figure*}[tp]
\centering
\cardtitle{Phase~1 Pre-Labeling Prompt \textnormal{\small(shared by GPT-5, Gemini~2.5~Pro, Doubao-Seed-1.6-Vision, GLM-4.5V)}}\\[5pt]
\prosecard{%
\small
\ptag{TASK INSTRUCTION} You are an expert Image Quality Assessment (IQA) analyst specialising in nighttime photography. Analyse the provided photograph and output the requested JSON fields. Your analysis must strictly assess the six distortion types listed in the \texttt{distortion\_severity} field. \emph{Motion Blur}: blur from moving objects while the background stays relatively sharp. \emph{Camera Shake}: blur affecting the entire image due to camera movement during exposure.

\smallskip
\ptag{PRIORITY GUIDANCE} When selecting \texttt{primary\_distortion}, apply a two-step process. (1)~\emph{Irreversibility}: prioritise distortions that cause irreversible loss of crucial information (e.g.\ blur or severe exposure issues) over inherent issues such as noise. (2)~\emph{Conflict resolution}: if several distortions are equally severe and irreversible, select the one most affecting the primary subject, or the one that is visually most dominant.

\smallskip
\ptag{SEVERITY DEFINITIONS} \emph{Severe}: prominent, severely degrades quality. \emph{Moderate}: noticeable, clearly degrades quality. \emph{Slight}: negligible or absent.

\smallskip
\ptag{OUTPUT FORMAT} A single, raw, syntactically correct JSON object with the following nine fields:
}\\[6pt]
\rule{\textwidth}{0.4pt}\\[5pt]
\codecard{\nightjsonbox}\\[5pt]
\botrule
\caption{Pre-labeling prompt template shared by all four frontier MLLMs in the Phase~1 ensemble. Each model receives this identical prompt alongside the target nighttime image and returns a structured JSON annotation following the schema shown at the bottom.}
\label{fig:annotation_prompt}
\end{figure*}

\begin{lrbox}{\nightjsonexbox}
\begin{minipage}{\textwidth}
\ttfamily\footnotesize
\begin{verbatim}
{
  "scene_category": "Outdoor urban street view",
  "distortion_severity": {
      "Noise": "Moderate",        "Underexposure": "Severe",
      "Overexposure": "Moderate", "Motion_Blur": "Slight",
      "Camera_Shake": "Slight",   "Focus_Blur": "Slight"
  },
  "primary_distortion": "Underexposure",
  "intended_subject": "White sedan parked on a dark street
      in front of brightly lit storefronts.",
  "subject_distortions": {
      "Noise": "Moderate", "Underexposure": "Severe"
  },
  "scene_description": "Nighttime urban street; white sedan
      on dark asphalt (foreground); sidewalk, tree, stacked
      chairs (midground); lit restaurant facade with neon
      trim and a circular illuminated logo (background).",
  "distortion_analysis": {
      "Underexposure": "Global - car, street, unlit
          storefronts, and darker building areas.",
      "Overexposure": "Local - neon signage, logo, awning.",
      "Noise": "Global - mid-tone and dark regions."
  },
  "content_impact_discussion": "Underexposure hides
      foreground detail and the car's plate/interior; noise
      adds grain in dark regions; local overexposure clips
      neon highlights.",
  "explanation": [
      "Underexposure is primary: irreversible loss of
          crucial detail across the scene.",
      "Overexposure is local, limited to signage."
  ]
}
\end{verbatim}
\end{minipage}
\end{lrbox}

\begin{figure*}[tp]
\centering
\cardtitle{Annotation Example \textnormal{\small(NightIQA-Real; ensemble output verified by human annotators)}}\\[5pt]
\codecard{\nightjsonexbox}\\[5pt]
\botrule
\caption{A complete annotation example from NightIQA-Real, produced by the multi-model ensemble and verified by human annotators (long free-text values reflowed for display).}
\label{fig:annotation_example}
\end{figure*}

\section{Scene-Level Deduplication and Train/Benchmark Partitioning}
\label{app:dedup}

We enforce scene-level separation through a four-stage pipeline: (1)~privacy-aware face filtering, (2)~DINOv2 feature extraction, (3)~agglomerative clustering with cluster-level partitioning, and (4)~a manual review that removes any residual visual overlap between the two splits.

\paragraph{DINOv2 Feature Extraction.}
Each image is embedded using \textbf{DINOv2 ViT-B/14}~\cite{supp-oquab2023dinov2}. Images are resized via letterbox to $518{\times}518$ with bicubic interpolation and ImageNet normalization. The \texttt{[CLS]} token serves as the scene descriptor, $\ell_2$-normalized as:
\begin{equation}
    \hat{\mathbf{f}} = \frac{\mathbf{f}}{\|\mathbf{f}\|_2},\quad \hat{\mathbf{f}} \in\mathbb{R}^{768}.
\end{equation}

\paragraph{Agglomerative Scene Clustering.}
Given the normalized feature matrix $\hat{F}\in\mathbb{R}^{N\times 768}$ ($N{=}1{,}769$), we apply hierarchical clustering with cosine distance $d(\hat{\mathbf{f}}_i, \hat{\mathbf{f}}_j) = 1 - \hat{\mathbf{f}}_i^\top \hat{\mathbf{f}}_j$, complete linkage, and threshold $\tau{=}0.45$, yielding \textbf{334 scene clusters}.

\paragraph{Cluster-Level Split.}
Clusters are sorted by ascending size and greedily assigned to training until reaching the target count. Since partitioning operates at the cluster level, no scene overlap or near-duplicate leakage occurs. This yields the 953-image GRPO training corpus and the 816-image held-out benchmark source, from which the 1{,}078 NightIQA-Bench questions are drawn. Table~\ref{tab:split_summary} summarizes the partition.

\paragraph{Manual Overlap Review.}
After the automatic cluster-level assignment, an annotator inspects boundary cases across the two splits and removes any residual near-duplicate scenes, so the training corpus and the benchmark source share no visually similar scenes.

\begin{table}[htbp]
    \centering
    \caption{Scene-level train/benchmark partition statistics. All splits are performed at the cluster level to prevent scene leakage.}
    \label{tab:split_summary}
    \renewcommand{\arraystretch}{1.05}
    \begin{tabular}{lc}
        \toprule
        \textbf{Subset} & \textbf{Images} \\
        \midrule
        GRPO training corpus (small clusters)     & 953 \\
        Benchmark source (remaining clusters)     & 816 \\
        \midrule
        Total                               & 1{,}769 \\
        \bottomrule
    \end{tabular}
\end{table}

\section{NightIQA-Bench: Detailed Examples and Statistics}
\label{app:bench}

\subsection{Question Distribution and Filtering}
\label{app:bench_stats}

The final validated benchmark contains \textbf{1{,}078} questions, distributed across the four cognitive levels as L1: 99, L2: 344, L3: 344, and L4: 291. Questions are first generated from the NightIQA-Real annotations and then admitted only after passing the two-stage filter introduced in Section~\ref{sec:benchmark}, which we expand here into four concrete steps:

\begin{enumerate}[noitemsep, leftmargin=1.6em, label=\textbf{(\arabic*)}]
    \item \textbf{Candidate generation.} For each image, multiple-choice questions spanning all four cognitive levels are generated jointly with chain-of-thought rationales grounded in the structured annotations; Level~2 yields two question types (primary-distortion identification and severity grading).
    \item \textbf{Structural screening.} Level completeness is enforced, and each rationale is checked for forbidden leakage patterns that would reveal the answer.
    \item \textbf{MLLM-as-a-judge filter} (first filter stage). An automatic judge screens candidates for logical errors, weak distractors, and insufficient discriminability.
    \item \textbf{Domain-expert review} (second filter stage). Human experts verify question clarity and visual grounding, retaining only questions that pass both filter stages.
\end{enumerate}

\noindent The resulting MCQ+CoT pairs serve as the source for downstream GRPO training.

\subsection{MCQ Structure and CoT Generation Protocol}
\label{app:bench_protocol}

To keep the benchmark consistent across images and graders, every question is generated against a fixed schema rather than free-form prompting. We detail the three components that govern this schema below.

\paragraph{Controlled vocabulary.}
All questions and rationales are restricted to the annotation vocabulary defined in Section~\ref{sec:real}: the six canonical distortion types (\textit{Noise}, \textit{Underexposure}, \textit{Overexposure}, \textit{Motion Blur}, \textit{Focus Blur}, \textit{Camera Shake}), the severity levels, and the spatial-scope labels (\textit{global} / \textit{local}). Binding the wording to this taxonomy ensures that questions probe the same attributes that the annotations record, and that an answer can always be traced back to a grounded field.

\paragraph{Cognitive levels and option design.}
Each image yields questions across all four cognitive levels, with Level~2 contributing two complementary question types, so a single image produces both a distortion-identification item and a severity-grading item at that level. The levels follow the body's definitions: \textbf{L1 (Perception)} recognizes the subject and scene content visible under nighttime conditions; \textbf{L2 (Technical Assessment)} identifies the primary distortion and grades its severity; \textbf{L3 (Localization)} probes spatial scope and the affected region; and \textbf{L4 (Reasoning)} probes causal and comparative judgments that require integrating multiple cues. Most questions are four-option single-answer items following a \textbf{1+2+1} design: one correct answer, two hard distractors drawn from attributes that are plausible for the same scene (e.g., a neighboring severity level, a co-occurring distortion type, or an alternative affected region), and one easy distractor that is clearly inconsistent with the image. The severity-grading items at Level~2 instead use the three fixed options \textit{Slight} / \textit{Moderate} / \textit{Severe}, so that any severity level can be the correct answer. This composition keeps the task discriminative while avoiding trivially guessable items.

\paragraph{Blind-test CoT protocol.}
The chain-of-thought rationale attached to each question is generated under a blind-test constraint: the rationale must describe the visual evidence and reasoning that lead to the answer \emph{without} stating the option letter or copying the answer text, and a fixed list of forbidden answer-revealing phrases is enforced during structural screening (Section~\ref{app:bench_stats}, step~2). This prevents the rationale from leaking the answer and forces the model to articulate grounded perceptual evidence, which is what the description reward $r_{\text{desc}}$ (Section~\ref{app:reward}) later supervises.

\subsection{Extended Benchmark Examples}
\label{app:bench_examples}

\begin{figure*}[tp]
    \centering
    \benchpanel{Level 1: Perception (99 questions)}{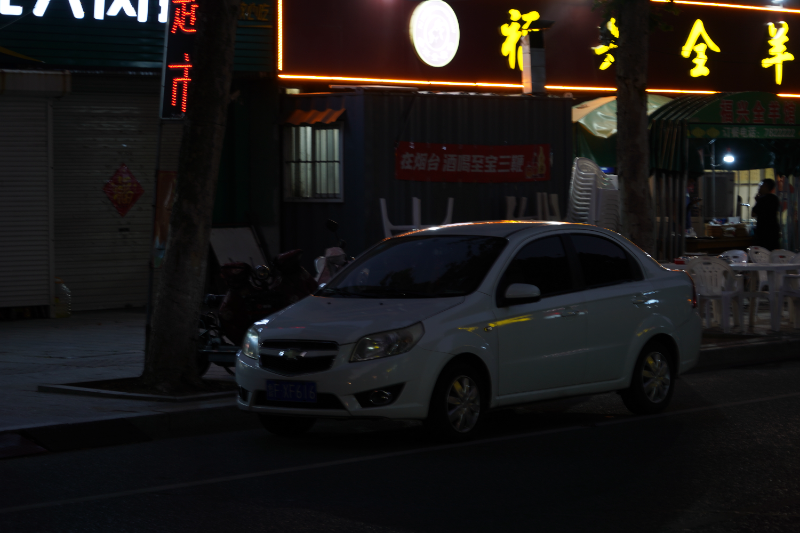}{%
        \bquestion{Which description best matches the visible scene content?}
        \begin{boptions}
            \item A nighttime urban street with very dark, heavily out-of-focus foreground foliage and a prominent blurred pink/red flower cluster near the \emph{center-left}, a curving road, and a scooter headlight in the background.\correct
            \item Identical scene, but the flower cluster sits near the \emph{center-right}.
            \item Identical layout, but the cluster is \emph{purple} instead of pink/red.
            \item A person riding a scooter with its headlight on is visible on the roadway in the background.
        \end{boptions}
        \bcot{The scene reads as a dark nighttime street with a large, very soft (out-of-focus) foreground plant mass and a pink/red cluster near the center-left blocking part of the view; the road curves right and a single bright headlight suggests a scooter in the distance. Option~B shifts the cluster to center-right, which does not match. Option~C keeps the layout but changes the color away from pink/red. Option~D is true but only names a background element, ignoring the dominant foreground foliage that defines the scene.}}

    \vspace{12pt}

    \benchpanel{Level 2: Technical Assessment (344 questions)}{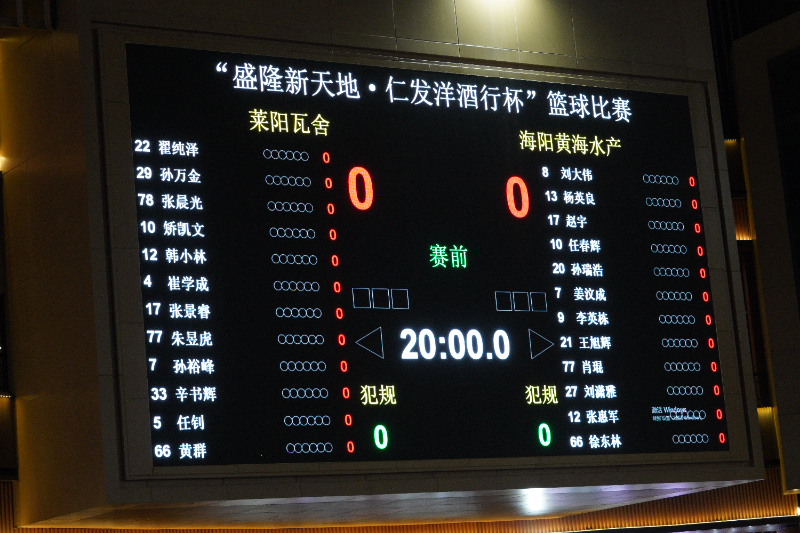}{%
        \bquestion{Which distortion is the most significant technical issue affecting the image?}
        \begin{boptions}
            \item Moderate Underexposure in the foreground grass and shadow regions under the right-side awning that hides texture and turns details into dark silhouettes.
            \item Moderate Overexposure around the upper-left streetlight and nearby highlights that clips bright detail and creates glare drawing attention away from the scene.\correct
            \item \emph{Slight} Overexposure around the same upper-left streetlight region.
            \item Slight Noise visible across the darker sky and shadowed areas.
        \end{boptions}
        \bcot{The brightest failure is the intense streetlight in the upper-left, where highlights are clipped and glare is obvious, making Moderate Overexposure the dominant issue. One near-miss shifts priority to foreground Underexposure, which is present but less visually dominant than the blown streetlight. Another keeps the same region but understates it as Slight, when the clipping and halo indicate stronger impact. Noise exists in dark areas but is minor and not the degradation asked for.}}

    \caption{Benchmark examples for cognitive levels L1--L2. Each panel pairs the nighttime image with its multiple-choice question, options (correct answer marked), and the jointly generated chain-of-thought rationale.}
    \label{fig:bench_l1l2}
\end{figure*}

\begin{figure*}[tp]
    \centering
    \benchpanel{Level 3: Localization (344 questions)}{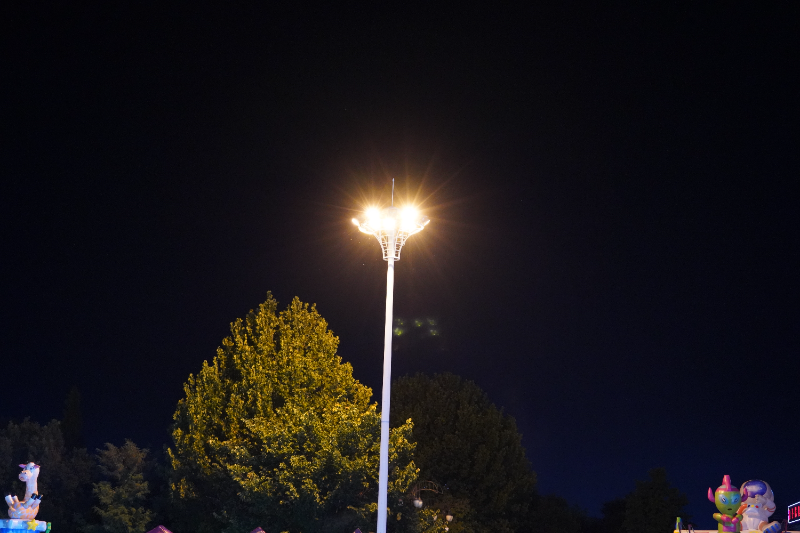}{%
        \bquestion{Where is the Overexposure most concentrated in the scene?}
        \begin{boptions}
            \item On the left tree canopy, especially the bright green leaves nearest the streetlight.
            \item On and immediately around the streetlight's lamp heads at the top of the pole, forming a bright halo and starburst that hides lamp detail.\correct
            \item On the upper half of the pole below the lamp heads, where the pole surface is uniformly blown out.
            \item Across the dark night sky in the background, where the image looks mottled rather than smooth.
        \end{boptions}
        \bcot{The brightest clipped region is clearly the lamp cluster at the top, with a halo/starburst that overwhelms the lamp structure. The left tree is illuminated but still shows leaf texture rather than the same concentrated clipping. The pole below the lamps is bright but not uniformly blown out like the lamp heads. The sky issue is real but corresponds to Noise, not Overexposure.}}

    \vspace{12pt}

    \benchpanel{Level 4: Reasoning (291 questions)}{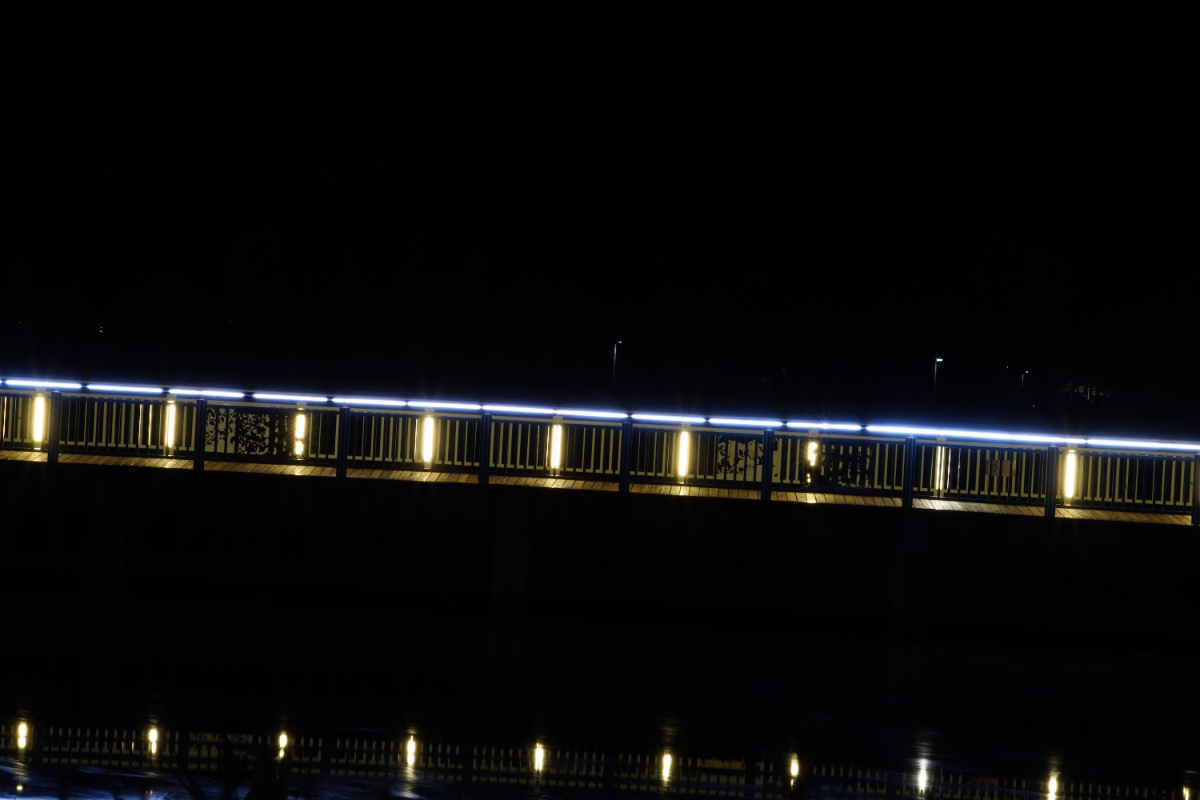}{%
        \bquestion{Why are the bridge's structural details and surrounding context hard to discern in this image?}
        \begin{boptions}
            \item Moderate Overexposure is the primary cause, clipping most tones first, with Severe Underexposure only secondarily reducing shadow detail.
            \item Severe Underexposure is the primary cause, crushing most of the frame into near-black, with Moderate Overexposure secondarily clipping the bright blue LED handrail and lamps.\correct
            \item Moderate Overexposure is primary, clipping the LED handrail and lamps first, with Severe Underexposure only secondarily crushing the rest.
            \item The bridge runs horizontally over dark water that reflects the blue handrail and warm vertical lights.
        \end{boptions}
        \bcot{Details disappear chiefly because most regions (sky, water, and much of the bridge structure) are pushed into near-black, consistent with Severe Underexposure removing midtones and shadows; on top of that, the brightest elements show Moderate Overexposure that clips highlights. Options A and C reverse the causal priority, wrongly treating Overexposure as the main driver. Option D truthfully describes composition and reflections but does not explain the degradation mechanism.}}

    \caption{Benchmark examples for cognitive levels L3--L4 (continued from Figure~\ref{fig:bench_l1l2}).}
    \label{fig:bench_l3l4}
\end{figure*}

Figures~\ref{fig:bench_l1l2} and~\ref{fig:bench_l3l4} present full-width benchmark views for all four cognitive levels.

\section{NightIQA-Syn: Degradation Pipeline Details}
\label{app:syn}

Each distortion is synthesized at two severity levels (\textit{Medium} and \textit{Severe}). Medium parameters are stochastically sampled; Severe parameters are derived by applying a fixed scaling factor.

\paragraph{Note on severity levels.}
NightIQA-Real and NightIQA-Syn use different severity granularities by design. Real captures carry the three perceptual levels used by human annotators (\textit{Slight} / \textit{Moderate} / \textit{Severe}, Section~\ref{app:annotation_schema}), because real degradations form a continuum that annotators can resolve into three bands. Synthetic degradations are generated at two controllable levels (\textit{Medium} / \textit{Severe}), since the parameters are programmatically set and the goal is to expose the model to a clearly moderate and a clearly strong instance of each type. When the two sources are combined for training, the synthetic \textit{Medium} level is aligned with the real \textit{Slight}--\textit{Moderate} range and \textit{Severe} with \textit{Severe}, so that severity supervision remains consistent across the real and synthetic portions of the data.

\subsection{Global-Level Degradation Algorithms}
\label{app:global_deg}

\paragraph{Noise.}
A black-level offset ($+2$) is added, then \textbf{chromatic sensor noise} is injected via \texttt{ISONoise} (color shift $c \in [0.07, 0.09]$, intensity $\alpha \in [0.32, 0.37]$), followed by \textbf{additive Gaussian grain} ($\mu{=}3$, $\sigma_g \in [4.8, 5.3]$). Severe: all parameters $\times 1.4$ (capped at 1.0).

\paragraph{Underexposure.}
With probability $p \in [0.2, 0.4]$, low-intensity \texttt{ISONoise} ($c \in [0.01, 0.03]$, $\alpha \in [0.05, 0.10]$) is injected, then the image is multiplied by brightness factor $\beta \in [0.35, 0.55]$. Severe: $\beta / 2.5$, noise params ${\times} 2.5$.

\paragraph{Global Overexposure.}
In CIE~LAB space: a highlight mask is created by thresholding $L$ at $\tau \in [180, 200]$ and smoothing with Gaussian radius $r \in [71, 81]$\,px; gamma correction ($\gamma \in [0.50, 0.55]$) and brightness addition ($\Delta L \in [60, 65]$) are applied; the bloom mask is composited additively. Severe: $\gamma / 1.3$, $\Delta L {\times} 1.3$, $\tau {\times} 0.7$, $r {\times} 1.3$.

\paragraph{Camera Shake.}
\texttt{MotionBlur} with kernel extent $k \in [40, 60]$\,px and random angle $\phi \in [0, 360]^{\circ}$. Severe: $k {\times} 2.0$.

\begin{figure*}[tp]
    \centering
    \begin{subfigure}{0.49\textwidth}
        \centering
        \includegraphics[width=\textwidth]{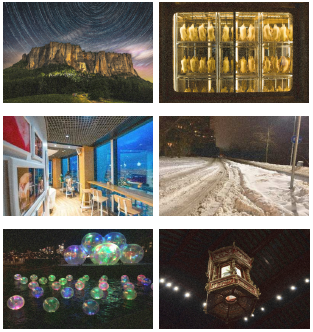}
        \caption{Noise}
        \label{fig:global_noise}
    \end{subfigure}
    \hfill
    \begin{subfigure}{0.49\textwidth}
        \centering
        \includegraphics[width=\textwidth]{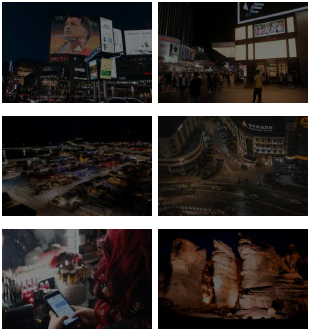}
        \caption{Underexposure}
        \label{fig:global_underexposure}
    \end{subfigure}

    \vspace{6pt}

    \begin{subfigure}{0.49\textwidth}
        \centering
        \includegraphics[width=\textwidth]{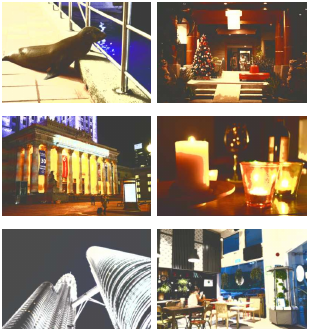}
        \caption{Global Overexposure}
        \label{fig:global_overexposure}
    \end{subfigure}
    \hfill
    \begin{subfigure}{0.49\textwidth}
        \centering
        \includegraphics[width=\textwidth]{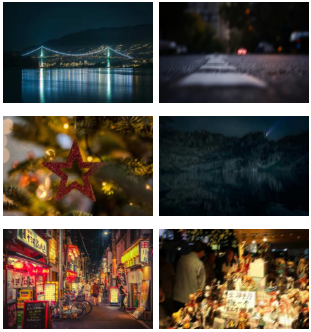}
        \caption{Camera Shake}
        \label{fig:global_camerashake}
    \end{subfigure}
    \caption{Visual examples of global-level degradations in NightIQA-Syn, grouped by distortion type. Within each category, the columns show two representative scenes and the rows correspond to increasing severity.}
    \label{fig:global_distortion}
\end{figure*}

Figure~\ref{fig:global_distortion} shows representative examples of each global-level distortion at both severity levels, and Table~\ref{tab:global_params} summarizes all parameter ranges.

\begin{table}[htbp]
    \centering
    \caption{Parameter configurations for global-level degradation operators in NightIQA-Syn.}
    \label{tab:global_params}
    \renewcommand{\arraystretch}{1.08}
    \scriptsize
    \setlength{\tabcolsep}{3.5pt}
    \begin{tabular}{@{}llcc@{}}
        \toprule
        \textbf{Distortion} & \textbf{Parameter} & \textbf{Medium} & \textbf{Severe} \\
        \midrule
        \multirow{3}{*}{Noise}
            & Color shift $c$                       & $[0.07,\; 0.09]$      & $c_m \times 1.4$ \\
            & ISO intensity $\alpha$                & $[0.32,\; 0.37]$      & $\alpha_m \times 1.4$ \\
            & Gaussian sigma $\sigma_g$             & $[4.8,\; 5.3]$        & $\sigma_{g,m} \times 1.4$ \\
        \midrule
        \multirow{4}{*}{Underexposure}
            & Brightness factor $\beta$             & $[0.35,\; 0.55]$      & $\beta_m / 2.5$ \\
            & Noise probability $p$                 & $[0.2,\; 0.4]$        & $p_m \times 2.5$ \\
            & Noise color shift $c$                 & $[0.01,\; 0.03]$      & $c_m \times 2.5$ \\
            & Noise intensity $\alpha$              & $[0.05,\; 0.10]$      & $\alpha_m \times 2.5$ \\
        \midrule
        \multirow{4}{*}{Global Overexposure}
            & Gamma $\gamma$                        & $[0.50,\; 0.55]$      & $\gamma_m / 1.3$ \\
            & Brightness addition $\Delta L$        & $[60,\; 65]$          & $\Delta L_m \times 1.3$ \\
            & Bloom threshold $\tau$                & $[180,\; 200]$        & $\tau_m \times 0.7$ \\
            & Bloom radius $r$ (px)                 & $[71,\; 81]$          & $r_m \times 1.3$ \\
        \midrule
        \multirow{2}{*}{Camera Shake}
            & Kernel extent $k$ (px)                & $[40,\; 60]$          & $k_m \times 2.0$ \\
            & Motion angle $\phi$ (deg)             & $[0,\; 360]$          & same as Medium \\
        \bottomrule
    \end{tabular}
\end{table}

\subsection{Instance-Level Degradation Algorithms}
\label{app:instance_deg}

\paragraph{Target Identification (Stage~1).}
Before any instance-level operator is applied, a vision-language model inventories the salient scene elements that can serve as degradation targets (Section~\ref{sec:synthetic}): movable objects (\textit{person}, \textit{vehicle}, \textit{animal}) for Motion Blur, point light sources (\textit{neon\_sign}, \textit{headlight}, \textit{taillight}, \textit{lamp}, \textit{spotlight}, etc.) for Local Overexposure, and depth-separated objects for Focus Blur. The model is queried with the prompt of Figure~\ref{fig:target_id_prompt} and returns up to five categories from a fixed content vocabulary; only images that yield valid targets enter the corresponding rendering branch, after which Rex-Omni and SAM3 ground the named targets into pixel-accurate masks (Stage~2).

\begin{lrbox}{\nighttargetbox}
\begin{minipage}{\textwidth}
\ttfamily\footnotesize
\begin{verbatim}
person, vehicle, animal, storefront, sign, billboard,
light, illuminated_billboard, neon_sign, led_display,
screen, traffic_light, light_strip, string_lights, lamp,
chandelier, spotlight, taillight, monitor,
illuminated_window, headlight, sky, sun, cloud, tree,
door, escalator, elevator, table, chair, sofa, bed,
shelf, shadow
\end{verbatim}
\end{minipage}
\end{lrbox}

\begin{figure*}[tp]
\centering
\cardtitle{Instance-Level Target-Identification Prompt \textnormal{\small(Stage~1)}}\\[5pt]
\prosecard{%
\small
\ptag{TASK} You are an image tagger. From the category vocabulary below, choose up to five categories that are most confidently visible in the image. If fewer than five categories are clearly visible, return fewer.

\smallskip
\ptag{OUTPUT FORMAT} Return only a single JSON object: \texttt{\{"img\_path": "<path>", "tag": ["tag1", ..., "tag5"]\}}.
}\\[6pt]
\rule{\textwidth}{0.4pt}\\[5pt]
\codecard{\nighttargetbox}\\[5pt]
\botrule
\caption{Stage~1 target-identification prompt for the instance-level pipeline. The model selects up to five categories from the fixed content vocabulary (bottom). Movable objects, point light sources, and depth-separated objects serve as candidate targets for Motion Blur, Local Overexposure, and Focus Blur respectively; only images with valid targets enter the matching rendering branch.}
\label{fig:target_id_prompt}
\end{figure*}

\paragraph{Local Overexposure.}
Light-source masks are combined via bitwise OR into $M$. The light layer $L = I \odot M$ is extracted, a bloom layer is generated via Gaussian filtering ($\sigma \in [45, 50]$), and composited: $O = I + 0.5L + \alpha \cdot \mathrm{Bloom}(L)$ with $\alpha \in [1.6, 1.7]$. Severe: $\sigma {\times} 2.6$, $\alpha {\times} 2.6$.

\paragraph{Motion Blur.}
Object masks are loaded, softened, and combined. An adaptive displacement $d$ is derived from the bounding-box diagonal scaled by $\alpha \in [0.10, 0.15]$. The image is linearized (gamma 2.2), then $N \in [30, 35]$ warped frames are accumulated along a randomized trajectory with vertical shift $v \in [0.008, 0.012]$, followed by matched noise injection ($\eta \in [0.01, 0.015]$) and soft compositing. Severe: all parameters $\times 3.5$.

\paragraph{Focus Blur.}
A monocular depth map guides layered disk-kernel convolution. The focal depth is derived from the secondary-object mask; the blur radius per pixel is $r = r_{\max} \cdot |d - d_f| / \max|d - d_f|$ with $r_{\max} \in [32, 38]$\,px, divided into $K{=}5$ layers. Severe: $r_{\max} {\times} 2.3$.

\begin{figure*}[tp]
    \centering
    \begin{subfigure}{0.32\textwidth}
        \centering
        \includegraphics[width=\textwidth]{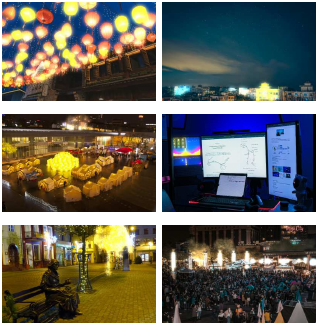}
        \caption{Local Overexposure}
        \label{fig:local_overexp}
    \end{subfigure}
    \hfill
    \begin{subfigure}{0.32\textwidth}
        \centering
        \includegraphics[width=\textwidth]{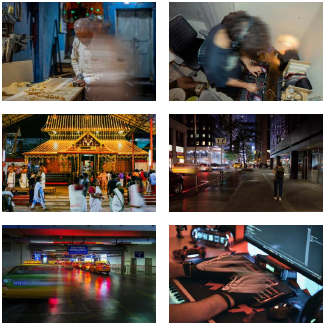}
        \caption{Motion Blur}
        \label{fig:local_motion}
    \end{subfigure}
    \hfill
    \begin{subfigure}{0.32\textwidth}
        \centering
        \includegraphics[width=\textwidth]{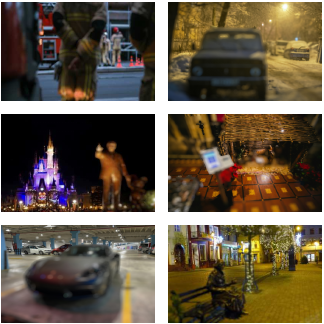}
        \caption{Focus Blur}
        \label{fig:local_focus}
    \end{subfigure}
    \caption{Visual examples of instance-level degradations in NightIQA-Syn, grouped by distortion type. Within each category, the columns show two representative scenes and the rows correspond to increasing severity.}
    \label{fig:local_distortion}
\end{figure*}

Figure~\ref{fig:local_distortion} shows representative examples of each instance-level distortion, and Table~\ref{tab:instance_params} summarizes all parameter ranges.

\begin{table}[htbp]
    \centering
    \caption{Parameter configurations for instance-level degradation operators in NightIQA-Syn.}
    \label{tab:instance_params}
    \renewcommand{\arraystretch}{1.08}
    \scriptsize
    \setlength{\tabcolsep}{1.6pt}
    \begin{tabular}{@{}llcc@{}}
        \toprule
        \textbf{Distortion} & \textbf{Parameter} & \textbf{Medium} & \textbf{Severe} \\
        \midrule
        \multirow{2}{*}{Local Overexposure}
            & Bloom sigma $\sigma$ (px)             & $[45,\; 50]$           & $\sigma_m \times 2.6$ \\
            & Bloom intensity $\alpha$              & $[1.6,\; 1.7]$        & $\alpha_m \times 2.6$ \\
        \midrule
        \multirow{4}{*}{Motion Blur}
            & Displacement intensity $\alpha$       & $[0.10,\; 0.15]$      & $\alpha_m \times 3.5$ \\
            & Warp steps $N$                        & $[30,\; 35]$          & $N_m \times 3.5$ \\
            & Vertical shift factor $v$             & $[0.008,\; 0.012]$    & $v_m \times 3.5$ \\
            & Matched noise level $\eta$            & $[0.01,\; 0.015]$     & $\eta_m \times 3.5$ \\
        \midrule
        \multirow{2}{*}{Focus Blur}
            & Max blur radius $r_{\max}$ (px)       & $[32,\; 38]$          & $r_{\max,m} \times 2.3$ \\
            & Depth layers $K$                      & \multicolumn{2}{c}{5 (fixed)} \\
        \bottomrule
    \end{tabular}
\end{table}

\subsection{Core Implementation of the Degradation Operators}
\label{app:syn_code}

This subsection lists the core routines of our released preprocessing toolkit, condensed from the production scripts (logging, I/O, multi-threading, and SFT-text generation removed) to expose the degradation logic itself. Every operator exposes a \texttt{MEDIUM\_PARAMS} dictionary and a scalar \texttt{SEVERE\_FACTOR}: the \emph{Medium} level samples parameters from the listed ranges, and the \emph{Severe} level scales them by \texttt{SEVERE\_FACTOR} (Section~\ref{app:syn}). The four \emph{global} operators in Figure~\ref{fig:code_global} act on the entire frame; the three \emph{instance-level} operators in Figures~\ref{fig:code_motion}--\ref{fig:code_local} act only within object masks or a depth map. We use \texttt{cv2} (OpenCV), \texttt{np} (NumPy), \texttt{A} (Albumentations), and \texttt{torch}/\texttt{F} (PyTorch / \texttt{torch.nn.functional}), with \texttt{DEVICE} the active CUDA device.

\begin{lrbox}{\codeboxglobal}
\begin{minipage}{\textwidth}
\scriptsize
\begin{verbatim}
# ---------- ISO sensor noise (global) ----------
SEVERE_FACTOR = 1.4
MEDIUM_PARAMS = {"color_shift": (0.07, 0.09),
                 "intensity":   (0.32, 0.37),
                 "gauss_sigma": (4.8, 5.3)}

def iso_noise(bgr, color, intensity, g_sigma):
    rgb = cv2.cvtColor(bgr, cv2.COLOR_BGR2RGB)
    rgb = cv2.add(rgb, np.array([2, 2, 2], np.uint8))   # lift black level
    rgb = A.ISONoise(color_shift=(color, color),
                     intensity=(intensity, intensity), p=1.0)(image=rgb)["image"]
    noise = np.random.normal(3, g_sigma, rgb.shape).astype(np.float32)
    rgb = np.clip(rgb.astype(np.float32) + noise, 0, 255).astype(np.uint8)
    return cv2.cvtColor(rgb, cv2.COLOR_RGB2BGR)

# ---------- Underexposure (global) ----------
SEVERE_FACTOR = 2.5
MEDIUM_PARAMS = {"brightness_factor": (0.35, 0.55),
                 "noise_probability": (0.2, 0.4)}

def underexpose(bgr, brightness_factor, noise_prob, iso_color, iso_intensity):
    if np.random.rand() < noise_prob:                   # probabilistic sensor noise
        rgb = cv2.cvtColor(bgr, cv2.COLOR_BGR2RGB)
        rgb = A.ISONoise(color_shift=(iso_color, iso_color),
                         intensity=(iso_intensity, iso_intensity))(image=rgb)["image"]
        bgr = cv2.cvtColor(rgb, cv2.COLOR_RGB2BGR)
    out = bgr.astype(np.float32) / 255.0 * brightness_factor   # linear dimming
    return np.clip(out * 255, 0, 255).astype(np.uint8)

# ---------- Global overexposure / bloom ----------
SEVERE_FACTOR = 1.3
MEDIUM_PARAMS = {"gamma": (0.5, 0.55), "brightness_add": (60, 65),
                 "bloom_threshold": (180, 200), "bloom_radius": (71, 81)}

def overexpose(bgr, gamma, b_add, t_val, r_val):
    l, a, b = cv2.split(cv2.cvtColor(bgr, cv2.COLOR_BGR2LAB))
    _, sharp = cv2.threshold(l, t_val, 255, cv2.THRESH_BINARY)
    r_val = r_val if r_val % 2 else r_val + 1            # odd kernel size
    bloom = cv2.GaussianBlur(sharp, (r_val, r_val), 0)
    lut = np.array([((i / 255.0) ** (1.0 / gamma)) * 255
                    for i in range(256)]).astype(np.uint8)
    l = cv2.add(cv2.LUT(l, lut), np.array([b_add], np.uint8))
    l = cv2.add(l, bloom)
    return cv2.cvtColor(cv2.merge([l, a, b]), cv2.COLOR_LAB2BGR)

# ---------- Camera shake (whole-image motion blur) ----------
SEVERE_FACTOR = 2.0
MEDIUM_PARAMS = {"blur_limit": (40, 60)}

def camera_shake(bgr, limit, angle):
    rgb = cv2.cvtColor(bgr, cv2.COLOR_BGR2RGB)
    t = A.MotionBlur(blur_limit=(limit, limit),
                     angle_range=(angle, angle), allow_shifted=True, p=1.0)
    return cv2.cvtColor(t(image=rgb)["image"], cv2.COLOR_RGB2BGR)
\end{verbatim}
\end{minipage}
\end{lrbox}

\begin{figure*}[tp]
\centering
\cardtitle{Global Degradation Operators \textnormal{\small(noise, underexposure, overexposure, camera shake)}}\\[5pt]
\codecard{\codeboxglobal}\\[5pt]
\botrule
\caption{Core routines for the four global operators. Each is applied to the full frame; \texttt{Severe} multiplies the sampled \texttt{MEDIUM\_PARAMS} by \texttt{SEVERE\_FACTOR}.}
\label{fig:code_global}
\end{figure*}

\begin{lrbox}{\codeboxmotion}
\begin{minipage}{\textwidth}
\scriptsize
\begin{verbatim}
# ---------- Motion blur (instance-level, physical optical integration) ----------
SEVERE_FACTOR = 3.5
MEDIUM_PARAMS = {"steps": (30, 35), "intensity": (0.1, 0.15),
                 "vertical_shift_factor": (0.008, 0.012),
                 "noise_range": (0.01, 0.015)}

def adaptive_magnitude(w, h, intensity, level):
    diag = np.sqrt(w**2 + h**2)
    mag  = diag * intensity
    max_limit = {"slight": 25.0, "medium": 55.0}.get(level, 200.0)
    if diag < 200:                                  # damp small objects
        mag *= max(0.4, diag / 200.0)
    if mag > max_limit:                             # soft ceiling
        mag = max_limit + (mag - max_limit) * 0.25
    mag = max(1.5, mag)
    ratio = 0.75 if level == "severe" else 0.5      # cap by object size
    return min(mag, min(w, h) * ratio)

def motion_blur_torch(img, mask, steps, trans_x, trans_y, noise=0.05):
    x = torch.from_numpy(img[..., ::-1].copy()).float()
    x = x.permute(2, 0, 1).unsqueeze(0).to(DEVICE) / 255.0
    m = torch.from_numpy(mask).float()[None, None].to(DEVICE)
    x = torch.pow(x + 1e-6, 2.2)                    # sRGB -> linear light
    B, C, H, W = x.shape
    yy, xx = torch.meshgrid(torch.linspace(-1, 1, H, device=DEVICE),
                            torch.linspace(-1, 1, W, device=DEVICE), indexing='ij')
    base = torch.stack((xx, yy), -1)[None]
    acc_img = torch.zeros_like(x); acc_m = torch.zeros_like(m); wsum = 0.0
    for i in range(steps):                          # integrate along the path
        g = base.clone()
        g[..., 0] -= (trans_x / steps * i) * 2 / W
        g[..., 1] -= (trans_y / steps * i) * 2 / H
        wi = F.grid_sample(x, g, align_corners=True, padding_mode="border")
        wm = F.grid_sample(m, g, align_corners=True, padding_mode="zeros")
        acc_img += wi * wm; acc_m += wm; wsum += 1.0
    acc_m = torch.clamp(acc_m, min=1e-3)
    obj = torch.pow(acc_img / acc_m + 1e-6, 1 / 2.2)
    obj = torch.clamp(obj + torch.randn_like(obj) * noise, 0, 1)   # matched grain
    fg  = torch.clamp(acc_m / (wsum + 1e-6), 0, 1)
    out = obj * fg + torch.pow(x + 1e-6, 1 / 2.2) * (1 - fg)        # composite
    out = out.squeeze(0).permute(1, 2, 0).cpu().numpy()
    return np.clip(out * 255, 0, 255).astype(np.uint8)[..., ::-1]
\end{verbatim}
\end{minipage}
\end{lrbox}

\begin{figure*}[tp]
\centering
\cardtitle{Instance-Level Motion Blur \textnormal{\small(physical optical-flow integration, GPU)}}\\[5pt]
\codecard{\codeboxmotion}\\[5pt]
\botrule
\caption{Motion blur is rendered in linear light by integrating the masked object along its displacement path, then alpha-composited back over the sharp background; displacement is set adaptively from object size.}
\label{fig:code_motion}
\end{figure*}

\begin{lrbox}{\codeboxfocus}
\begin{minipage}{\textwidth}
\scriptsize
\begin{verbatim}
# ---------- Focus blur (instance-level, depth-aware layered defocus) ----------
SEVERE_FACTOR = 2.3
MEDIUM_PARAMS = {"max_radius": (32, 38)}

def disk_kernel(radius):
    ks = int(radius * 2 + 1)
    yy, xx = torch.meshgrid(torch.linspace(-radius, radius, ks),
                            torch.linspace(-radius, radius, ks), indexing='ij')
    k = torch.clamp(radius - torch.sqrt(xx**2 + yy**2) + 0.5, 0, 1)
    return (k / k.sum()).to(DEVICE)                 # normalized disk (CoC)

def layered_blur(img, depth, focal_depth, max_radius, layers=5):
    d = F.avg_pool2d(depth, 5, 1, 2)
    diff = torch.abs(d - focal_depth)
    target = diff / (diff.max() + 1e-6) * max_radius   # blur radius per pixel
    step = max_radius / (layers - 1)
    out = img.clone()
    for i in range(layers - 1):                     # discretize depth into layers
        lo, hi = i * step, (i + 1) * step
        m = (target >= lo) & (target < hi)
        if not torch.any(m): continue
        k = disk_kernel((lo + hi) / 2)
        pad = k.shape[0] // 2
        k4 = k[None, None].repeat(3, 1, 1, 1)
        b = F.conv2d(F.pad(img, (pad,) * 4, mode='reflect'), k4, groups=3)
        out[m.repeat(1, 3, 1, 1)] = b[m.repeat(1, 3, 1, 1)]
    return out

def focus_blur_torch(img, depth, focal_point, max_radius):
    x = torch.from_numpy(img[..., ::-1].copy()).float()
    x = x.permute(2, 0, 1).unsqueeze(0).to(DEVICE) / 255.0
    x = torch.pow(x + 1e-6, 2.2)                    # sRGB -> linear light
    d = torch.from_numpy(depth.astype(np.float32))[None, None].to(DEVICE)
    fx, fy = focal_point
    focal = torch.median(d[0, 0, max(0, fy-2):fy+3, max(0, fx-2):fx+3]).item()
    out = layered_blur(x, d, focal, max_radius, layers=5)
    out = torch.pow(out + 1e-6, 1 / 2.2).squeeze(0).permute(1, 2, 0).cpu().numpy()
    return np.clip(out * 255, 0, 255).astype(np.uint8)[..., ::-1]
\end{verbatim}
\end{minipage}
\end{lrbox}

\begin{figure*}[tp]
\centering
\cardtitle{Instance-Level Focus Blur \textnormal{\small(depth-aware layered defocus, GPU)}}\\[5pt]
\codecard{\codeboxfocus}\\[5pt]
\botrule
\caption{Defocus is applied in depth-ordered layers: each pixel's circle-of-confusion radius grows with its distance from the focal plane, and a normalized disk kernel is convolved per layer in linear light.}
\label{fig:code_focus}
\end{figure*}

\begin{lrbox}{\codeboxlocal}
\begin{minipage}{\textwidth}
\scriptsize
\begin{verbatim}
# ---------- Local overexposure (instance-level atmospheric bloom) ----------
SEVERE_FACTOR = 2.6
MEDIUM_PARAMS = {"sigma": (45.0, 50.0), "intensity": (1.6, 1.7)}

def atmospheric_bloom(bgr, mask, sigma, intensity):
    img = bgr.astype(np.float32) / 255.0
    m = mask.astype(np.float32) / 255.0
    if m.ndim == 2:
        m = cv2.merge([m, m, m])
    light = img * m                                 # isolate light sources
    bloom = cv2.GaussianBlur(light, (0, 0), sigmaX=sigma, sigmaY=sigma)
    out = img + light * 0.5 + bloom * intensity     # additive optical blend
    return (np.clip(out, 0, 1.0) * 255).astype(np.uint8)
\end{verbatim}
\end{minipage}
\end{lrbox}

\begin{figure*}[tp]
\centering
\cardtitle{Instance-Level Local Overexposure \textnormal{\small(mask-driven atmospheric bloom)}}\\[5pt]
\codecard{\codeboxlocal}\\[5pt]
\botrule
\caption{Local overexposure isolates masked light sources, generates a wide Gaussian halo, and blends it additively, reproducing the bloom around bright nighttime sources.}
\label{fig:code_local}
\end{figure*}

\subsection{Synthetic Distortion Validation}
\label{app:syn_validation}

Before SFT sampling (\suppref{app:sft_sampling}), every generated image passes an automated quality-control step that verifies the intended distortion was actually rendered and is consistent with its structured label. An MLLM inspector receives the degraded image together with the expected distortion type, severity, scope, and the structured chain-of-thought description, and is queried with the prompt of Figure~\ref{fig:validation_prompt}. The judge returns a JSON verdict covering distortion presence, type match, scope accuracy, target reasonableness (for instance-level distortions), and severity consistency, plus an objective \textit{perceived-severity} rating assessed independently of the label and a $1$--$10$ quality score. An image is retained only when all boolean checks pass, the quality score is $\ge 7$, and no critical issue is flagged; otherwise it is discarded. The check is scope-aware: global distortions must affect the entire frame uniformly, whereas instance-level distortions must be confined to a semantically appropriate target (light sources for Local Overexposure, moving objects for Motion Blur, depth-separated objects for Focus Blur), so that a mismatched target (e.g., motion blur on a static building) is marked as a failure.

\begin{lrbox}{\nightvalbox}
\begin{minipage}{\textwidth}
\ttfamily\footnotesize
\begin{verbatim}
{
  "distortion_present":      true|false,
  "distortion_matches_type": true|false,
  "scope_accurate":          true|false,
  "target_identified":       true|false,   # local only
  "target_reasonable":       true|false,   # local only
  "target_matches_sft":      true|false,   # local only
  "severity_appropriate":    true|false,
  "perceived_severity":      "slight|medium|severe",
  "severity_label_accurate": true|false,
  "sft_description_accurate":true|false,
  "overall_quality_score":   1-10,
  "validation_passed":       true|false,
  "issues_found":            ["...", ...],
  "reasoning":               "brief explanation"
}
\end{verbatim}
\end{minipage}
\end{lrbox}

\begin{figure*}[tp]
\centering
\cardtitle{Synthetic Distortion Validation Prompt \textnormal{\small(MLLM-as-inspector)}}\\[5pt]
\prosecard{%
\small
\ptag{TASK} You are a professional image quality inspector. Validate whether this nighttime photograph exhibits the specific distortion described in the training data. You are given the expected distortion type, severity, scope, key visual features, and the structured chain-of-thought description (scene, scope, distortion specifics, target, labeled severity).

\smallskip
\ptag{GLOBAL CHECKS} Confirm that the distortion (i)~affects the \emph{entire} frame uniformly, (ii)~matches the expected type, (iii)~is consistent with the labeled severity, and (iv)~exhibits the expected visual features; localized effects under a Global label are marked as failures.

\smallskip
\ptag{LOCAL CHECKS} Confirm that the distortion is confined to a specific target and that the target is \emph{semantically appropriate}: light sources for Local Overexposure, moving objects for Motion Blur, and depth-separated objects for Focus Blur. The observed target must match the one stated in the description, and non-target regions must remain relatively unaffected; an unreasonable target (e.g., motion blur on a static building) is marked as a failure.

\smallskip
\ptag{SEVERITY} Independently assess \emph{perceived severity} (slight/medium/severe) from the image alone, then check whether it matches the labeled level.

\smallskip
\ptag{DECISION RULE} \texttt{validation\_passed} is true only if all boolean checks pass, \texttt{overall\_quality\_score} $\ge 7$, and \texttt{issues\_found} is empty.

\smallskip
\ptag{OUTPUT FORMAT} Return only the JSON object below (no extra text):
}\\[6pt]
\rule{\textwidth}{0.4pt}\\[5pt]
\codecard{\nightvalbox}\\[5pt]
\botrule
\caption{Synthetic distortion validation prompt. Each generated NightIQA-Syn image is inspected by an MLLM judge that returns the structured verdict (bottom); only images that pass every boolean check with a quality score $\ge 7$ enter the SFT sampling pool (\suppref{app:sft_sampling}).}
\label{fig:validation_prompt}
\end{figure*}

\subsection{SFT Training Data Sampling Strategy}
\label{app:sft_sampling}

From the 80{,}000-image synthetic corpus, \textbf{6{,}769} images are sampled via stratified sampling along distortion type and severity level, with near-balanced coverage where validated samples permit. Table~\ref{tab:sft_dist} reports the distribution.

\begin{table}[htbp]
    \centering
    \caption{SFT training data distribution by distortion type and severity.}
    \label{tab:sft_dist}
    \renewcommand{\arraystretch}{1.08}
    \begin{tabular}{lccr}
        \toprule
        \textbf{Distortion Type} & \textbf{Medium} & \textbf{Severe} & \textbf{Total} \\
        \midrule
        Noise               & 500 & 500 & 1{,}000 \\
        Underexposure       & 500 & 500 & 1{,}000 \\
        Global Overexposure & 500 & 500 & 1{,}000 \\
        Local Overexposure  & 501 & 268 & 769 \\
        Camera Shake        & 500 & 500 & 1{,}000 \\
        Motion Blur         & 500 & 500 & 1{,}000 \\
        Focus Blur          & 500 & 500 & 1{,}000 \\
        \midrule
        Total               & 3{,}501 & 3{,}268 & 6{,}769 \\
        \bottomrule
    \end{tabular}
\end{table}

\section{Supervised Fine-Tuning: Full Details}
\label{app:sft}

\subsection{SFT Data Construction}
\label{app:sft_data}

Each degraded image $\tilde{x}$ is annotated with degradation type $y^{\text{type}}$, severity $y^{\text{sev}}$ (Medium/Severe), and scope $y^{\text{scope}} \in \{\text{global}, \text{local}\}$. Supervision is constructed in two formats: \textbf{Basic QA} (three single-turn pairs probing type, severity, and scope) and \textbf{Chain-of-Thought} (a structured response sequentially describing content, distortion, and spatial distribution).

\subsection{SFT Training Objective}
\label{app:sft_objective}

The SFT stage uses the masked next-token prediction loss, restricted to response tokens:
\begin{equation}
    \mathcal{L}_{\mathrm{SFT}} = - \sum_{t=1}^{T} m_t \log p_{\theta}(z_t \mid z_{<t}),
    \label{eq:sft_loss_supp}
\end{equation}
where $m_t \in \{0,1\}$ masks out image and prompt tokens. The supervision implicitly targets:
\begin{equation}
    \hat{y} = \{ \hat{y}^{\text{type}},\; \hat{y}^{\text{sev}},\; \hat{y}^{\text{scope}} \}.
\end{equation}

The learning rate follows cosine annealing with linear warmup. During warmup ($0 \le s < S_w$), $\eta_s = \eta_0 \, s / S_w$; during annealing ($S_w \le s \le S$),
\begin{equation}
    \eta_s = \eta_{\min} + \tfrac{1}{2}(\eta_0 - \eta_{\min})\bigl(1 + \cos(\pi\,\phi_s)\bigr),
    \quad \phi_s = \tfrac{s - S_w}{S - S_w},
    \label{eq:cosine_lr}
\end{equation}
where $S_w = 0.01 \cdot S$ is the warmup length.

\subsection{SFT Hyperparameters}
\label{app:sft_hyper}

Table~\ref{tab:sft_hyper} summarizes the SFT hyperparameters.

\begin{table}[htbp]
    \centering
    \caption{Hyperparameters for the SFT stage.}
    \label{tab:sft_hyper}
    \renewcommand{\arraystretch}{1.08}
    \begin{tabular}{ll}
        \toprule
        \textbf{Hyperparameter} & \textbf{Value} \\
        \midrule
        Base model                  & Qwen3-VL-8B \\
        Fine-tuning strategy        & Full-parameter SFT \\
        Training epochs             & 3 \\
        Learning rate $\eta_0$      & $1 \times 10^{-5}$ \\
        LR schedule                 & Cosine annealing \\
        Warmup ratio                & 0.01 \\
        Optimizer                   & AdamW \\
        Data precision              & bfloat16 \\
        \bottomrule
    \end{tabular}
\end{table}

\section{GRPO: Full Formulation and Training Details}
\label{app:grpo}

\subsection{Group Relative Policy Optimization}
\label{app:grpo_formulation}

For each query $q$ with image $x$, GRPO samples $G$ outputs $\{o_i\}_{i=1}^{G}$ from $\pi_{\theta_{\text{old}}}$, scores them to obtain $\{r_i\}$, and computes the group-relative advantage:
\begin{equation}
    \hat{A}_i = \frac{r_i - \mathrm{mean}(\{r_j\})}{\mathrm{std}(\{r_j\})}.
    \label{eq:advantage}
\end{equation}
The policy is updated via:
\begin{equation}
    \begin{aligned}
        \mathcal{J}_{\text{GRPO}}(\theta)
        &= \mathbb{E}\Big[ \tfrac{1}{G} \sum_{i=1}^{G}
            \min\!\big(\rho_i\hat{A}_i, \\
        &\qquad \mathrm{clip}(\rho_i,1{-}\epsilon,1{+}\epsilon)\hat{A}_i\big) \\
        &\qquad {}-\beta\,\mathbb{D}_{\mathrm{KL}}
            [\pi_\theta\,\|\,\pi_{\text{ref}}]\Big],
    \end{aligned}
    \label{eq:grpo_obj}
\end{equation}
where $\rho_i = \pi_{\theta}(o_i \mid q) / \pi_{\theta_{\text{old}}}(o_i \mid q)$. The per-token KL divergence is:
\begin{equation}
    \begin{split}
        \mathbb{D}_{\mathrm{KL}}\!\left[\pi_{\theta} \,\|\, \pi_{\text{ref}}\right] = \frac{1}{|o_i|} \sum_{t=1}^{|o_i|} \Big[ \frac{\pi_{\theta}(o_{i,t} \mid q, o_{i,<t})}{\pi_{\text{ref}}(o_{i,t} \mid q, o_{i,<t})} \\
        {}- \log \frac{\pi_{\theta}(o_{i,t} \mid q, o_{i,<t})}{\pi_{\text{ref}}(o_{i,t} \mid q, o_{i,<t})} - 1 \Big].
    \end{split}
    \label{eq:kl}
\end{equation}

\subsection{Structured Output Format}
\label{app:output_format}

The model generates responses in the following format:

{\footnotesize
\begin{verbatim}
<thinking>
[Analysis of visual attributes and
 quality reasoning process]
</thinking>
<answer>[Option letter]</answer>
\end{verbatim}
}

\subsection{Composite Reward Function}
\label{app:reward}

The total reward is:
\begin{equation}
    r_{\text{total}} = w_{\text{acc}} \cdot r_{\text{acc}} + w_{\text{desc}} \cdot r_{\text{desc}} + w_{\text{fmt}} \cdot r_{\text{fmt}},
    \label{eq:reward}
\end{equation}
with $w_{\text{acc}} {=} 1.0$, $w_{\text{desc}} {=} 0.8$, $w_{\text{fmt}} {=} 0.5$.

\paragraph{Answer Accuracy $r_{\text{acc}}$.}
$r_{\text{acc}} = \mathbb{1}[\hat{y}_{\text{ans}} = y_{\text{ans}}^{*}]$.

\paragraph{Description Quality $r_{\text{desc}}$.}
\begin{equation}
    r_{\text{desc}} = \frac{1}{2}\!\left(\mathrm{BLEU\text{-}4}(\hat{d}, d^{*}) + \mathrm{ROUGE\text{-}L}(\hat{d}, d^{*})\right),
\end{equation}
where $\hat{d}$ is the \texttt{<thinking>} text and $d^{*}$ is the reference description. BLEU-4 measures $n$-gram precision with brevity penalty:
\begin{equation}
    \mathrm{BLEU\text{-}4}(\hat{d}, d^{*}) = \mathrm{BP} \cdot \exp\!\left(\frac{1}{4}\sum_{n=1}^{4} \log p_n\right),
\end{equation}
and ROUGE-L is based on the longest common subsequence:
\begin{equation}
    \mathrm{ROUGE\text{-}L}(\hat{d}, d^{*}) = \frac{2 \cdot P_{\text{lcs}} \cdot R_{\text{lcs}}}{P_{\text{lcs}} + R_{\text{lcs}}},
\end{equation}
with $P_{\text{lcs}} = |\mathrm{LCS}(\hat{d}, d^{*})| / |\hat{d}|$ and $R_{\text{lcs}} = |\mathrm{LCS}(\hat{d}, d^{*})| / |d^{*}|$.

\paragraph{Format Compliance $r_{\text{fmt}}$.}
\begin{equation}
    r_{\text{fmt}} = \mathbb{1}\!\left[\,\text{output matches } \mathcal{P}_{\text{fmt}}\,\right],
    \label{eq:reward_fmt}
\end{equation}
where the pattern $\mathcal{P}_{\text{fmt}}$ requires \texttt{<thinking>.*</thinking>}, optional whitespace, then \texttt{<answer>.*</answer>}.

\paragraph{Weight Rationale.}
$w_{\text{acc}}$ is highest (sparse binary signal needs strong gradient drive); $w_{\text{desc}}$ is slightly lower (continuous signal already provides dense gradients); $w_{\text{fmt}}$ is lowest (format compliance is largely learned during SFT).

\subsection{GRPO Hyperparameters}
\label{app:grpo_hyper}

Table~\ref{tab:grpo_hyper} summarizes the GRPO training configuration.

\begin{table}[htbp]
    \centering
    \caption{Hyperparameters for the GRPO alignment stage.}
    \label{tab:grpo_hyper}
    \renewcommand{\arraystretch}{1.08}
    \begin{tabular}{@{}l p{0.46\columnwidth}@{}}
        \toprule
        \textbf{Hyperparameter} & \textbf{Value} \\
        \midrule
        Base model (after SFT)          & Qwen3-VL-8B (SFT ckpt) \\
        Framework / Parallelism         & ms-swift / DeepSpeed~\cite{supp-rasley2020deepspeed} ZeRO-3 \\
        GPUs / Precision                & 6 / bfloat16 \\
        Training epochs                 & 5 \\
        Learning rate                   & $1 \times 10^{-6}$ \\
        Batch size (per device)         & 1 \\
        Gradient accumulation           & 4 \\
        Effective batch size            & $6 \times 1 \times 4 = 24$ \\
        Group size $G$ / Temperature    & 8 / 0.9 \\
        KL coefficient $\beta$          & 0.04 \\
        Max gen / input length          & 1{,}024 / 2{,}048 tokens \\
        Max input resolution            & 262{,}144 pixels \\
        Warmup ratio                    & 0.05 \\
        Selected epoch                  & 4 (best avg \& L4) \\
        \midrule
        \multicolumn{2}{@{}p{\columnwidth}@{}}{\textbf{Reward weights:} $w_{\text{acc}}{=}1.0$, $w_{\text{desc}}{=}0.8$, $w_{\text{fmt}}{=}0.5$} \\
        \bottomrule
    \end{tabular}
\end{table}

\subsection{System Prompt}
\label{app:system_prompt}

During GRPO training, the following system prompt is prepended:

\begin{quote}
\small
\textit{You are a nighttime image quality assessment expert. Begin your response with a \texttt{<thinking>} tag. Inside, provide a fluent, non-repetitive prose analysis of the nighttime image's visual characteristics, including brightness distribution, noise level, detail preservation, color fidelity, and any visible distortions. Do NOT use option prefixes such as ``A.'', ``B.'' inside the thinking section to avoid leaking your answer tendency. After the analysis, output the final option letter inside an \texttt{<answer>} tag.}
\end{quote}

\section{Ethical Considerations and Privacy}
\label{app:ethics}

\paragraph{Data Collection.}
All NightIQA-Real images were captured by the research team; no images were sourced from surveillance systems, social media, or unconsented third parties.

\paragraph{Privacy Protection.}
Every image underwent automated face detection followed by manual review for identifiable information. Images failing either check were excluded.

\paragraph{Licensing.}
NightIQA-Syn uses Unsplash~\cite{supp-unsplash2020dataset} source images (Unsplash License). NightIQA-Real and NightIQA-Bench will be released under the Creative Commons Attribution-NonCommercial-ShareAlike 4.0 (CC BY-NC-SA 4.0) license for research purposes, consistent with the non-commercial terms of the Unsplash source imagery.

\newpage
{\small
\fontencoding{T1}\selectfont
\urlstyle{same}

}

\begin{thebibliography}{53}
\providecommand{\natexlab}[1]{#1}
\providecommand{\url}[1]{\texttt{#1}}
\expandafter\ifx\csname urlstyle\endcsname\relax
  \providecommand{\doi}[1]{doi: #1}\else
  \providecommand{\doi}{doi: \begingroup \urlstyle{rm}\Url}\fi

\bibitem[{Anthropic}(2025)]{anthropic2025claudehaiku}
{Anthropic}.
\newblock Introducing {Claude} {Haiku} 4.5.
\newblock \url{https://www.anthropic.com/news/claude-haiku-4-5}, 2025.

\bibitem[Bai et~al.(2025)Bai, Cai, Chen, Chen, Chen, Cheng, Deng, Ding, Gao,
  Ge, et~al.]{bai2025qwen3}
Shuai Bai, Yuxuan Cai, Ruizhe Chen, Keqin Chen, Xionghui Chen, Zesen Cheng,
  Lianghao Deng, Wei Ding, Chang Gao, Chunjiang Ge, et~al.
\newblock {Qwen3-VL} technical report.
\newblock arXiv preprint arXiv:2511.21631, 2025.

\bibitem[Bhat et~al.(2023)Bhat, Birkl, Wofk, Wonka, and
  Müller]{bhat2023zoedepth}
Shariq~Farooq Bhat, Reiner Birkl, Diana Wofk, Peter Wonka, and Matthias
  Müller.
\newblock {ZoeDepth}: Zero-shot transfer by combining relative and metric
  depth.
\newblock arXiv preprint arXiv:2302.12288, 2023.

\bibitem[Bosse et~al.(2018)Bosse, Maniry, Müller, Wiegand, and
  Samek]{bosse2017deep}
Sebastian Bosse, Dominique Maniry, Klaus-Robert Müller, Thomas Wiegand, and
  Wojciech Samek.
\newblock Deep neural networks for no-reference and full-reference image
  quality assessment.
\newblock \emph{IEEE Transactions on Image Processing}, 27\penalty0
  (1):\penalty0 206--219, 2018.

\bibitem[Buslaev et~al.(2020)Buslaev, Iglovikov, Khvedchenya, Parinov,
  Druzhinin, and Kalinin]{buslaev2020albumentations}
Alexander Buslaev, Vladimir~I. Iglovikov, Eugene Khvedchenya, Alex Parinov,
  Mikhail Druzhinin, and Alexandr~A. Kalinin.
\newblock {Albumentations}: Fast and flexible image augmentations.
\newblock \emph{Information}, 11\penalty0 (2):\penalty0 125, 2020.

\bibitem[{ByteDance Seed}(2026)]{bytedance2026seed20mini}
{ByteDance Seed}.
\newblock {Seed2.0 Mini}.
\newblock \url{https://seed.bytedance.com/en/seed2}, 2026.

\bibitem[{ByteDance Seed Team}(2025)]{bytedance2025seed16vision}
{ByteDance Seed Team}.
\newblock {Introduction to Techniques Used in Seed1.6}.
\newblock
  \url{https://seed.bytedance.com/en/blog/introduction-to-techniques-used-in-seed1-6},
  2025.

\bibitem[Carion et~al.(2026)Carion, Gustafson, Hu, Debnath, Hu, Suris, Ryali,
  Alwala, Khedr, Huang, et~al.]{carion2025sam}
Nicolas Carion, Laura Gustafson, Yuan-Ting Hu, Shoubhik Debnath, Ronghang Hu,
  Didac Suris, Chaitanya Ryali, Kalyan~Vasudev Alwala, Haitham Khedr, Andrew
  Huang, et~al.
\newblock {SAM 3}: Segment anything with concepts.
\newblock arXiv preprint arXiv:2511.16719, 2026.

\bibitem[Chen et~al.(2024{\natexlab{a}})Chen, Mo, Hou, Wu, Liao, Sun, Yan, and
  Lin]{chen2024topiq}
Chaofeng Chen, Jiadi Mo, Jingwen Hou, Haoning Wu, Liang Liao, Wenxiu Sun, Qiong
  Yan, and Weisi Lin.
\newblock {TOPIQ}: A top-down approach from semantics to distortions for image
  quality assessment.
\newblock \emph{IEEE Transactions on Image Processing}, 33:\penalty0
  2404--2418, 2024{\natexlab{a}}.

\bibitem[Chen et~al.(2024{\natexlab{b}})Chen, Yang, Wu, Liao, Zhang, Wang, Sun,
  Yan, and Lin]{chen2024q}
Chaofeng Chen, Sensen Yang, Haoning Wu, Liang Liao, Zicheng Zhang, Annan Wang,
  Wenxiu Sun, Qiong Yan, and Weisi Lin.
\newblock {Q-Ground}: Image quality grounding with large multi-modality models.
\newblock In \emph{Proceedings of the 32nd ACM International Conference on
  Multimedia}, pages 486--495, 2024{\natexlab{b}}.

\bibitem[Chen et~al.(2024{\natexlab{c}})Chen, Wu, Wang, Su, Chen, Xing, Zhong,
  Zhang, Zhu, Lu, et~al.]{chen2024internvl}
Zhe Chen, Jiannan Wu, Wenhai Wang, Weijie Su, Guo Chen, Sen Xing, Muyan Zhong,
  Qinglong Zhang, Xizhou Zhu, Lewei Lu, et~al.
\newblock {InternVL}: Scaling up vision foundation models and aligning for
  generic visual-linguistic tasks.
\newblock In \emph{Proceedings of the IEEE/CVF Conference on Computer Vision
  and Pattern Recognition}, pages 24185--24198, 2024{\natexlab{c}}.

\bibitem[Comanici et~al.(2025)Comanici, Bieber, Schaekermann, Pasupat,
  Sachdeva, Dhillon, Blistein, Ram, Zhang, Rosen, et~al.]{comanici2025gemini}
Gheorghe Comanici, Eric Bieber, Mike Schaekermann, Ice Pasupat, Noveen
  Sachdeva, Inderjit Dhillon, Marcel Blistein, Ori Ram, Dan Zhang, Evan Rosen,
  et~al.
\newblock {Gemini} 2.5: Pushing the frontier with advanced reasoning,
  multimodality, long context, and next generation agentic capabilities.
\newblock arXiv preprint arXiv:2507.06261, 2025.

\bibitem[Fang et~al.(2020)Fang, Zhu, Zeng, Ma, and Wang]{fang2020perceptual}
Yuming Fang, Hanwei Zhu, Yan Zeng, Kede Ma, and Zhou Wang.
\newblock Perceptual quality assessment of smartphone photography.
\newblock In \emph{Proceedings of the IEEE/CVF Conference on Computer Vision
  and Pattern Recognition}, pages 3677--3686, 2020.

\bibitem[{GLM-V Team} et~al.(2026){GLM-V Team}, Hong, Yu, Gu, Wang, Gan, Tang,
  Cheng, Qi, Ji, et~al.]{hong2025glm45v}
{GLM-V Team}, Wenyi Hong, Wenmeng Yu, Xiaotao Gu, Guo Wang, Guobing Gan,
  Haomiao Tang, Jiale Cheng, Ji Qi, Junhui Ji, et~al.
\newblock {GLM-4.5V} and {GLM-4.1V-Thinking}: Towards versatile multimodal
  reasoning with scalable reinforcement learning.
\newblock arXiv preprint arXiv:2507.01006, 2026.

\bibitem[{Google DeepMind}(2026)]{google2025gemini31}
{Google DeepMind}.
\newblock {Gemini 3.1 Flash-Lite}.
\newblock
  \url{https://deepmind.google/models/model-cards/gemini-3-1-flash-lite/},
  2026.

\bibitem[Hosu et~al.(2020)Hosu, Lin, Sziranyi, and Saupe]{hosu2020koniq}
Vlad Hosu, Hanhe Lin, Tamas Sziranyi, and Dietmar Saupe.
\newblock {KonIQ-10k}: An ecologically valid database for deep learning of
  blind image quality assessment.
\newblock \emph{IEEE Transactions on Image Processing}, 29:\penalty0
  4041--4056, 2020.

\bibitem[Jiang et~al.(2025)Jiang, Huo, Chen, Xiong, Zeng, Chen, Ren, Yu, and
  Zhang]{jiang2025detect}
Qing Jiang, Junan Huo, Xingyu Chen, Yuda Xiong, Zhaoyang Zeng, Yihao Chen,
  Tianhe Ren, Junzhi Yu, and Lei Zhang.
\newblock Detect anything via next point prediction.
\newblock arXiv preprint arXiv:2510.12798, 2025.

\bibitem[Ke et~al.(2021)Ke, Wang, Wang, Milanfar, and Yang]{ke2021musiq}
Junjie Ke, Qifei Wang, Yilin Wang, Peyman Milanfar, and Feng Yang.
\newblock {MUSIQ}: Multi-scale image quality transformer.
\newblock In \emph{Proceedings of the IEEE/CVF International Conference on
  Computer Vision}, pages 5148--5157, 2021.

\bibitem[Li et~al.(2025)Li, Zhang, Zhao, Zhang, Li, Zhang, and Zhang]{li2025q}
Weiqi Li, Xuanyu Zhang, Shijie Zhao, Yabin Zhang, Junlin Li, Li Zhang, and Jian
  Zhang.
\newblock {Q-Insight}: Understanding image quality via visual reinforcement
  learning.
\newblock In \emph{Advances in Neural Information Processing Systems}, pages
  36802--36827, 2025.

\bibitem[Lin(2004)]{lin2004rouge}
Chin-Yew Lin.
\newblock {ROUGE}: A package for automatic evaluation of summaries.
\newblock In \emph{Text Summarization Branches Out}, pages 74--81, 2004.

\bibitem[Lin et~al.(2019)Lin, Hosu, and Saupe]{lin2019kadid}
Hanhe Lin, Vlad Hosu, and Dietmar Saupe.
\newblock {KADID-10k}: A large-scale artificially distorted {IQA} database.
\newblock In \emph{2019 Eleventh International Conference on Quality of
  Multimedia Experience (QoMEX)}, pages 1--3, 2019.

\bibitem[Liu et~al.(2024)Liu, Li, Li, and Lee]{liu2024improved}
Haotian Liu, Chunyuan Li, Yuheng Li, and Yong~Jae Lee.
\newblock Improved baselines with visual instruction tuning.
\newblock In \emph{Proceedings of the IEEE/CVF Conference on Computer Vision
  and Pattern Recognition}, pages 26296--26306, 2024.

\bibitem[Mittal et~al.(2012)Mittal, Moorthy, and Bovik]{mittal2012no}
Anish Mittal, Anush~Krishna Moorthy, and Alan~Conrad Bovik.
\newblock No-reference image quality assessment in the spatial domain.
\newblock \emph{IEEE Transactions on Image Processing}, 21\penalty0
  (12):\penalty0 4695--4708, 2012.

\bibitem[Mittal et~al.(2013)Mittal, Soundararajan, and Bovik]{mittal2012making}
Anish Mittal, Rajiv Soundararajan, and Alan~C. Bovik.
\newblock Making a “completely blind” image quality analyzer.
\newblock \emph{IEEE Signal Processing Letters}, 20\penalty0 (3):\penalty0
  209--212, 2013.

\bibitem[{OpenAI} et~al.(2024){OpenAI}, Hurst, Lerer, Goucher, Perelman,
  Ramesh, Clark, Ostrow, Welihinda, Hayes, et~al.]{hurst2024gpt}
{OpenAI}, Aaron Hurst, Adam Lerer, Adam~P. Goucher, Adam Perelman, Aditya
  Ramesh, Aidan Clark, AJ Ostrow, Akila Welihinda, Alan Hayes, et~al.
\newblock {GPT-4o} system card.
\newblock arXiv preprint arXiv:2410.21276, 2024.

\bibitem[Oquab et~al.(2024)Oquab, Darcet, Moutakanni, Vo, Szafraniec, Khalidov,
  Fernandez, Haziza, Massa, El-Nouby, et~al.]{oquab2023dinov2}
Maxime Oquab, Timoth{\'e}e Darcet, Th{\'e}o Moutakanni, Huy~V. Vo, Marc
  Szafraniec, Vasil Khalidov, Pierre Fernandez, Daniel Haziza, Francisco Massa,
  Alaaeldin El-Nouby, et~al.
\newblock {DINO}v2: Learning robust visual features without supervision.
\newblock \emph{Transactions on Machine Learning Research}, 2024.
\newblock Featured Certification.

\bibitem[Papineni et~al.(2002)Papineni, Roukos, Ward, and
  Zhu]{papineni2002bleu}
Kishore Papineni, Salim Roukos, Todd Ward, and Wei-Jing Zhu.
\newblock {B}leu: a method for automatic evaluation of machine translation.
\newblock In \emph{Proceedings of the 40th Annual Meeting of the {A}ssociation
  for {C}omputational {L}inguistics}, pages 311--318, 2002.

\bibitem[Ponomarenko et~al.(2015)Ponomarenko, Jin, Ieremeiev, Lukin,
  Egiazarian, Astola, Vozel, Chehdi, Carli, Battisti,
  et~al.]{ponomarenko2015image}
Nikolay Ponomarenko, Lina Jin, Oleg Ieremeiev, Vladimir Lukin, Karen
  Egiazarian, Jaakko Astola, Benoit Vozel, Kacem Chehdi, Marco Carli, Federica
  Battisti, et~al.
\newblock Image database {TID2013}: Peculiarities, results and perspectives.
\newblock \emph{Signal Processing: Image Communication}, 30:\penalty0 57--77,
  2015.

\bibitem[{Qwen Team}(2026)]{team2026qwen3}
{Qwen Team}.
\newblock {Qwen3.5}: Towards native multimodal agents.
\newblock \url{https://qwen.ai/blog?id=qwen3.5}, 2026.

\bibitem[Shao et~al.(2024)Shao, Wang, Zhu, Xu, Song, Bi, Zhang, Zhang, Li, Wu,
  et~al.]{shao2024deepseekmath}
Zhihong Shao, Peiyi Wang, Qihao Zhu, Runxin Xu, Junxiao Song, Xiao Bi, Haowei
  Zhang, Mingchuan Zhang, Y.~K. Li, Y. Wu, et~al.
\newblock {DeepSeekMath}: Pushing the limits of mathematical reasoning in open
  language models.
\newblock arXiv preprint arXiv:2402.03300, 2024.

\bibitem[Singh et~al.(2026)Singh, Fry, Perelman, Tart, Ganesh, El-Kishky,
  McLaughlin, Low, Ostrow, Ananthram, et~al.]{singh2025openai}
Aaditya Singh, Adam Fry, Adam Perelman, Adam Tart, Adi Ganesh, Ahmed El-Kishky,
  Aidan McLaughlin, Aiden Low, AJ Ostrow, Akhila Ananthram, et~al.
\newblock {OpenAI} {GPT-5} system card.
\newblock arXiv preprint arXiv:2601.03267, 2026.

\bibitem[Su et~al.(2020)Su, Yan, Zhu, Zhang, Ge, Sun, and Zhang]{su2020blindly}
Shaolin Su, Qingsen Yan, Yu Zhu, Cheng Zhang, Xin Ge, Jinqiu Sun, and Yanning
  Zhang.
\newblock Blindly assess image quality in the wild guided by a self-adaptive
  hyper network.
\newblock In \emph{Proceedings of the IEEE/CVF Conference on Computer Vision
  and Pattern Recognition}, pages 3667--3676, 2020.

\bibitem[{Unsplash}(2020)]{unsplash2020dataset}
{Unsplash}.
\newblock {Unsplash} dataset.
\newblock \url{https://unsplash.com/data}, 2020.

\bibitem[Wang et~al.(2023)Wang, Chan, and Loy]{wang2023exploring}
Jianyi Wang, Kelvin~C.K. Chan, and Chen~Change Loy.
\newblock Exploring {CLIP} for assessing the look and feel of images.
\newblock In \emph{Proceedings of the AAAI Conference on Artificial
  Intelligence}, pages 2555--2563, 2023.

\bibitem[Wang et~al.(2024)Wang, Bai, Tan, Wang, Fan, Bai, Chen, Liu, Wang, Ge,
  et~al.]{wang2024qwen2}
Peng Wang, Shuai Bai, Sinan Tan, Shijie Wang, Zhihao Fan, Jinze Bai, Keqin
  Chen, Xuejing Liu, Jialin Wang, Wenbin Ge, et~al.
\newblock {Qwen2-VL}: Enhancing vision-language model's perception of the world
  at any resolution.
\newblock arXiv preprint arXiv:2409.12191, 2024.

\bibitem[Wang et~al.(2025)Wang, Gao, Gu, Pu, Cui, Wei, Liu, Jing, Ye, Shao,
  et~al.]{wang2025internvl35}
Weiyun Wang, Zhangwei Gao, Lixin Gu, Hengjun Pu, Long Cui, Xingguang Wei,
  Zhaoyang Liu, Linglin Jing, Shenglong Ye, Jie Shao, et~al.
\newblock {InternVL3.5}: Advancing open-source multimodal models in
  versatility, reasoning, and efficiency.
\newblock arXiv preprint arXiv:2508.18265, 2025.

\bibitem[Wang et~al.(2004)Wang, Bovik, Sheikh, and Simoncelli]{wang2004image}
Zhou Wang, A.C. Bovik, H.R. Sheikh, and E.P. Simoncelli.
\newblock Image quality assessment: from error visibility to structural
  similarity.
\newblock \emph{IEEE Transactions on Image Processing}, 13\penalty0
  (4):\penalty0 600--612, 2004.

\bibitem[Wu et~al.(2024{\natexlab{a}})Wu, Zhang, Zhang, Chen, Liao, Wang, Li,
  Sun, Yan, Zhai, et~al.]{wu2024qbench}
Haoning Wu, Zicheng Zhang, Erli Zhang, Chaofeng Chen, Liang Liao, Annan Wang,
  Chunyi Li, Wenxiu Sun, Qiong Yan, Guangtao Zhai, et~al.
\newblock {Q-Bench}: A benchmark for general-purpose foundation models on
  low-level vision.
\newblock In \emph{International Conference on Learning Representations}, pages
  12547--12573, 2024{\natexlab{a}}.

\bibitem[Wu et~al.(2024{\natexlab{b}})Wu, Zhang, Zhang, Chen, Liao, Wang, Xu,
  Li, Hou, Zhai, et~al.]{wu2024q}
Haoning Wu, Zicheng Zhang, Erli Zhang, Chaofeng Chen, Liang Liao, Annan Wang,
  Kaixin Xu, Chunyi Li, Jingwen Hou, Guangtao Zhai, et~al.
\newblock {Q-Instruct}: Improving low-level visual abilities for multi-modality
  foundation models.
\newblock In \emph{Proceedings of the IEEE/CVF Conference on Computer Vision
  and Pattern Recognition}, pages 25490--25500, 2024{\natexlab{b}}.

\bibitem[Wu et~al.(2024{\natexlab{c}})Wu, Zhang, Zhang, Chen, Liao, Li, Gao,
  Wang, Zhang, Sun, et~al.]{wu2024qalign}
Haoning Wu, Zicheng Zhang, Weixia Zhang, Chaofeng Chen, Liang Liao, Chunyi Li,
  Yixuan Gao, Annan Wang, Erli Zhang, Wenxiu Sun, et~al.
\newblock {Q-Align}: Teaching {LMM}s for visual scoring via discrete
  text-defined levels.
\newblock In \emph{Proceedings of the 41st International Conference on Machine
  Learning}, pages 54015--54029, 2024{\natexlab{c}}.

\bibitem[Wu et~al.(2025{\natexlab{a}})Wu, Zhu, Zhang, Zhang, Chen, Liao, Li,
  Wang, Sun, Yan, et~al.]{wu2024towards}
Haoning Wu, Hanwei Zhu, Zicheng Zhang, Erli Zhang, Chaofeng Chen, Liang Liao,
  Chunyi Li, Annan Wang, Wenxiu Sun, Qiong Yan, et~al.
\newblock Towards open-ended visual quality comparison.
\newblock In \emph{Computer Vision -- ECCV 2024}, pages 360--377,
  2025{\natexlab{a}}.

\bibitem[Wu et~al.(2025{\natexlab{b}})Wu, Zou, Liang, Zhang, and
  Ma]{wu2025visualquality}
Tianhe Wu, Jian Zou, Jie Liang, Lei Zhang, and Kede Ma.
\newblock {VisualQuality-R1}: Reasoning-induced image quality assessment via
  reinforcement learning to rank.
\newblock In \emph{Advances in Neural Information Processing Systems}, pages
  88167--88190, 2025{\natexlab{b}}.

\bibitem[Xiang et~al.(2020)Xiang, Yang, and Guo]{xiang2019blind}
Tao Xiang, Ying Yang, and Shangwei Guo.
\newblock Blind night-time image quality assessment: Subjective and objective
  approaches.
\newblock \emph{IEEE Transactions on Multimedia}, 22\penalty0 (5):\penalty0
  1259--1272, 2020.

\bibitem[{Xiaomi MiMo Team}(2026{\natexlab{a}})]{xiaomi2026mimo}
{Xiaomi MiMo Team}.
\newblock Xiaomi {MiMo-V2-Omni}.
\newblock \url{https://mimo.xiaomi.com/mimo-v2-omni}, 2026{\natexlab{a}}.

\bibitem[{Xiaomi MiMo Team}(2026{\natexlab{b}})]{xiaomi2026mimov25}
{Xiaomi MiMo Team}.
\newblock Xiaomi {MiMo-V2.5}.
\newblock \url{https://mimo.xiaomi.com/mimo-v2-5}, 2026{\natexlab{b}}.

\bibitem[You et~al.(2025{\natexlab{a}})You, Cai, Gu, Xue, and
  Dong]{you2025teaching}
Zhiyuan You, Xin Cai, Jinjin Gu, Tianfan Xue, and Chao Dong.
\newblock Teaching large language models to regress accurate image quality
  scores using score distribution.
\newblock In \emph{Proceedings of the IEEE/CVF Conference on Computer Vision
  and Pattern Recognition}, pages 14483--14494, 2025{\natexlab{a}}.

\bibitem[You et~al.(2025{\natexlab{b}})You, Gu, Cai, Li, Zhu, Dong, and
  Xue]{you2025enhancing}
Zhiyuan You, Jinjin Gu, Xin Cai, Zheyuan Li, Kaiwen Zhu, Chao Dong, and Tianfan
  Xue.
\newblock Enhancing descriptive image quality assessment with a large-scale
  multi-modal dataset.
\newblock \emph{IEEE Transactions on Image Processing}, 34:\penalty0
  8201--8215, 2025{\natexlab{b}}.

\bibitem[You et~al.(2025{\natexlab{c}})You, Li, Gu, Yin, Xue, and
  Dong]{you2024depicting}
Zhiyuan You, Zheyuan Li, Jinjin Gu, Zhenfei Yin, Tianfan Xue, and Chao Dong.
\newblock Depicting beyond scores: Advancing image quality assessment through
  multi-modal language models.
\newblock In \emph{Computer Vision -- ECCV 2024}, pages 259--276,
  2025{\natexlab{c}}.

\bibitem[{Z.ai}(2025)]{zai2025glm46v}
{Z.ai}.
\newblock {GLM-4.6V}: Open source multimodal models with native tool use.
\newblock \url{https://z.ai/blog/glm-4.6v}, 2025.

\bibitem[Zhai and Min(2020)]{zhai2020perceptual}
Guangtao Zhai and Xiongkuo Min.
\newblock Perceptual image quality assessment: a survey.
\newblock \emph{Science China Information Sciences}, 63\penalty0 (11):\penalty0
  211301, 2020.

\bibitem[Zhang et~al.(2018)Zhang, Isola, Efros, Shechtman, and
  Wang]{zhang2018unreasonable}
Richard Zhang, Phillip Isola, Alexei~A. Efros, Eli Shechtman, and Oliver Wang.
\newblock The unreasonable effectiveness of deep features as a perceptual
  metric.
\newblock In \emph{Proceedings of the IEEE Conference on Computer Vision and
  Pattern Recognition}, pages 586--595, 2018.

\bibitem[Zhang et~al.(2023)Zhang, Zhai, Wei, Yang, and Ma]{zhang2023blind}
Weixia Zhang, Guangtao Zhai, Ying Wei, Xiaokang Yang, and Kede Ma.
\newblock Blind image quality assessment via vision-language correspondence: A
  multitask learning perspective.
\newblock In \emph{Proceedings of the IEEE/CVF Conference on Computer Vision
  and Pattern Recognition}, pages 14071--14081, 2023.

\bibitem[Zhang et~al.(2024)Zhang, Wu, Zhang, Zhai, and Lin]{zhang2024q}
Zicheng Zhang, Haoning Wu, Erli Zhang, Guangtao Zhai, and Weisi Lin.
\newblock {Q-Bench+}: A benchmark for multi-modal foundation models on
  low-level vision from single images to pairs.
\newblock \emph{IEEE Transactions on Pattern Analysis and Machine
  Intelligence}, 46\penalty0 (12):\penalty0 10404--10418, 2024.

\end{thebibliography}

\begin{thebibliography}{3}
\providecommand{\natexlab}[1]{#1}
\providecommand{\url}[1]{\texttt{#1}}
\expandafter\ifx\csname urlstyle\endcsname\relax
  \providecommand{\doi}[1]{doi: #1}\else
  \providecommand{\doi}{doi: \begingroup \urlstyle{rm}\Url}\fi

\bibitem[Oquab et~al.(2024)Oquab, Darcet, Moutakanni, Vo, Szafraniec, Khalidov,
  Fernandez, Haziza, Massa, El-Nouby, et~al.]{supp-oquab2023dinov2}
Maxime Oquab, Timoth{\'e}e Darcet, Th{\'e}o Moutakanni, Huy~V. Vo, Marc
  Szafraniec, Vasil Khalidov, Pierre Fernandez, Daniel Haziza, Francisco Massa,
  Alaaeldin El-Nouby, et~al.
\newblock {DINO}v2: Learning robust visual features without supervision.
\newblock \emph{Transactions on Machine Learning Research}, 2024.
\newblock Featured Certification.

\bibitem[Rasley et~al.(2020)Rasley, Rajbhandari, Ruwase, and
  He]{supp-rasley2020deepspeed}
Jeff Rasley, Samyam Rajbhandari, Olatunji Ruwase, and Yuxiong He.
\newblock {DeepSpeed}: System optimizations enable training deep learning
  models with over 100 billion parameters.
\newblock In \emph{Proceedings of the 26th ACM SIGKDD International Conference
  on Knowledge Discovery \& Data Mining}, pages 3505--3506, 2020.

\bibitem[{Unsplash}(2020)]{supp-unsplash2020dataset}
{Unsplash}.
\newblock {Unsplash} dataset.
\newblock \url{https://unsplash.com/data}, 2020.

\end{thebibliography}
\end{document}